\pdfoutput=1
\PassOptionsToPackage{table}{xcolor}
\documentclass[11pt]{article}

\usepackage[]{acl}

\usepackage{xcolor}
\usepackage{amssymb}
\usepackage{algorithm}
\usepackage{bbding}
\usepackage{pifont}
\usepackage{wasysym}
\usepackage{multirow}
\usepackage{algpseudocode} 
\usepackage{booktabs}

\usepackage{amsthm,multicol,enumerate}
\usepackage{makecell}
\usepackage{graphicx}
\usepackage{color,xcolor}
\usepackage{subcaption}
\usepackage{enumitem}
\usepackage{soul}
\usepackage{array}

\usepackage{array}
\usepackage{multirow}
\usepackage{soul}
\usepackage{times}
\usepackage{latexsym}
\usepackage{graphicx}
\usepackage{amsfonts}
\usepackage{adjustbox}
\usepackage{tabularx}
\usepackage{etoolbox}
\usepackage{pgf}
\usepackage{etoc}
\usepackage[T1]{fontenc}
\usepackage[utf8]{inputenc}

\usepackage{microtype}

\usepackage{inconsolata}

\title{Beyond Individual Facts: Investigating Categorical Knowledge Locality of Taxonomy and Meronomy Concepts in GPT Models}

\author{Christopher Burger \\
  Dept of Computer Science\\
  University of Mississippi\\
  \texttt{cburger@olemiss.edu} \\\And
  Yifan Hu \\
  Dept of Computer Science\\
  Northeastern University\\
  \texttt{yif.hu@northeastern.edu} \\\And
  Thai le \\
  Dept of Computer Science\\
  Indiana University\\
  \texttt{tle@iu.edu}
  }

\def\dataset{DARC}

\makeatletter
\newcommand\subsubsubsection{\@startsection{paragraph}{4}{\z@}{-2.5ex\@plus -1ex \@minus -.25ex}{1.25ex \@plus .25ex}{\normalfont\normalsize\bfseries}}
\newcommand\subsubsubsubsection{\@startsection{subparagraph}{5}{\z@}{-2.5ex\@plus -1ex \@minus -.25ex}{1.25ex \@plus .25ex}{\normalfont\normalsize\bfseries}}
\makeatother

\newcommand{\gradientcellM}[3]{
   \ifstrequal{#1}{1}{\cellcolor{#2!100.0}#1}
   {\ifstrequal{#1}{2}{\cellcolor{#2!92.0}#1}
   {\ifstrequal{#1}{3}{\cellcolor{#2!84.0}#1}
   {\ifstrequal{#1}{4}{\cellcolor{#2!76.0}#1}
   {\ifstrequal{#1}{5}{\cellcolor{#2!68.0}#1}
   {\ifstrequal{#1}{6}{\cellcolor{#2!60.0}#1}
   {\ifstrequal{#1}{7}{\cellcolor{#2!52.0}#1}
   {\ifstrequal{#1}{8}{\cellcolor{#2!44.0}#1}
   {\ifstrequal{#1}{9}{\cellcolor{#2!36.0}#1}
   {\ifstrequal{#1}{10}{\cellcolor{#2!28.0}#1}
   {\ifstrequal{#1}{11}{\cellcolor{#2!20.0}#1}
   {\ifstrequal{#1}{12}{\cellcolor{#2!12.0}#1}
   {\ifstrequal{#1}{13}{\cellcolor{#3!12.0}#1}
   {\ifstrequal{#1}{14}{\cellcolor{#3!20.0}#1}
   {\ifstrequal{#1}{15}{\cellcolor{#3!28.0}#1}
   {\ifstrequal{#1}{16}{\cellcolor{#3!36.0}#1}
   {\ifstrequal{#1}{17}{\cellcolor{#3!44.0}#1}
   {\ifstrequal{#1}{18}{\cellcolor{#3!52.0}#1}
   {\ifstrequal{#1}{19}{\cellcolor{#3!60.0}#1}
   {\ifstrequal{#1}{20}{\cellcolor{#3!68.0}#1}
   {\ifstrequal{#1}{21}{\cellcolor{#3!76.0}#1}
   {\ifstrequal{#1}{22}{\cellcolor{#3!84.0}#1}
   {\ifstrequal{#1}{23}{\cellcolor{#3!92.0}#1}
   {\ifstrequal{#1}{24}{\cellcolor{#3!100.0}#1}

   {}}}}}}}}}}}}}}}}}}}}}}}}

 }

\begin{document}
\maketitle
\begin{abstract}
The location of knowledge within Generative Pre-trained Transformer (GPT)-like models has seen extensive recent investigation. However, much of the work is focused towards determining locations of \textit{individual} facts, with the end goal being the editing of facts that are outdated, erroneous, or otherwise harmful, without the time and expense of retraining the entire model. In this work, we investigate a broader view of knowledge location, that of \textit{concepts or clusters of related information, instead of disparate individual facts}. To do this, we first curate a novel dataset, called \dataset, that includes a total of 34 concepts of $\sim$120K factual statements divided into two types of hierarchical categories, namely taxonomy and meronomy. Next, we utilize existing causal mediation analysis methods developed for determining regions of importance for individual facts and apply them to a series of related categories to provide detailed investigation into whether concepts are associated with distinct regions within these models. We find that related categories exhibit similar areas of importance in contrast to less similar categories. However, fine-grained localization of individual category subsets to specific regions is not apparent.

\end{abstract}

\section{Introduction}

The use of Large Language Models (LLMs) has exploded in cross-disciplinary use with GPT-like models at the forefront of this wave of popularity. With the ability to generate text with fluency indistinguishable from many humans to effective methods of linking, recalling, and summarizing text for information, it seems likely that LLMs will continue to encroach on previously human exclusive tasks. But as the effectiveness of LLMs continues to increase, so have the questions concerning \textit{how} LLMs make their decisions. And as LLMs become steadily more ingrained into society, the ability to explain \textit{why} one of these models has acted the way it did becomes of a question of substantial legal, social, and governmental importance. Just as human decision making is fallible, all models are subject to error. Being able to explain why an error was made not only allows developers a way to improve future models, but can provide important information to satisfy legal or ethical obligations should some harm result from a model's output.

The task of correcting erroneous output has become an important focus of recent research into how LLMs store information \cite{meng2022locating,vig2020causal,dai2022knowledge}, make inferences \cite{elazar2021amnesic,Feder_2021}, and from there how to modify the model's output without the appreciable time and expense of retraining \cite{mitchell2022fast,meng2022memit}. One of the successful approaches has been to apply techniques from causal analysis, which has been able to identify both specific locations for individual facts \cite{meng2022locating}, as well as locations for shared attributes, such as the idea of gender \cite{goyal2020explaining}. This approach is ideal from the perspective of explainability as casting model decisions in terms of causal processes provides inherent explainability. 
Assuming the causal analysis is accurate (a significant assumption), it directly reflects how the model operates.
Additionally, the standard visualization practice of modern causal analysis, a graph, is especially well suited to the structure of LLMs. LLM components can be viewed as the vertices with the edges being the possible pathways that exist from input to output. And since the causal graph is innately specified, using causal mediation analysis naturally follows.

Our goal is to more broadly explore the concept of locality in LLMs. Previous research by \cite{meng2022locating} demonstrated that specific instances of knowledge within these models can be traced to particular regions of importance. Building on this, we seek to investigate whether such regions exist for knowledge that is more extensive than individual facts. Specifically, we aim to determine if there are identifiable regions that encapsulate broader and more complex knowledge structures beyond isolated pieces of information. For this investigation, we focus on two common relational structures: First, \textit{taxonomy}, which organizes similar objects into successive groups; and second, \textit{meronomy}, which categorizes similar objects in part-whole relationships. We believe this investigation could provide deeper insights into how knowledge is organized and represented in LLMs.


Specifically, our investigation and contribution consists of:
\begin{enumerate}[leftmargin=\dimexpr\parindent-0.2\labelwidth\relax,itemsep=2pt]
   \item We extend the discussion of knowledge locality by investigating the plausibility of broader types of knowledge possessing locality within LLMs. We propose to study the knowledge locality in the context of \textit{intra-category} locality, to determines if related facts within a category are also located within a similar region in the structure of the model, and \textit{inter-category} locality, to examine whether knowledge in the similar categories are also close by the structure of the model.
   \item Our experiments provide evidence for both \textit{intra-cateogory} and \textit{inter-category} locality among hierarchical relations, especially related taxonomic groups. Meronomic relations also demonstrate weaker associations both \textit{intra-category} and \textit{inter-category}. 
   \item To accomplish the above, we created a novel data consisting of a series of related concepts to provide sufficient raw material to perform comparisons on, in contrast to existing datasets used in prior work that are themselves aggregates small amounts of many disparate ideas. The dataset comprises four major categories, encompassing 34 individual concepts and over 122,000 facts. It will be made available with the publication of the paper.
\end{enumerate}

\section{Related Work}
Determining where knowledge is located within LLMs has been a significant focus of recent research. Much of this work has centered around the task of updating outdated or otherwise erroneous information. As the initial pre-training from scratch of many transformers based models is now beyond the computational capacity for most practitioners and researchers, effective and robust editing tools are ideal maintain model correctness in light of constantly changing factual relations. Even for organizations with the resources to directly train these models, retraining to update a small set of facts would be inordinately impractical given the significant time and expense. 

Knowledge location has been an increasingly important subject in modern LLM research with a wide variety of hypothesis and methods \cite{roberts2020knowledge, jiang-etal-2020-know,hase2021language,dai2022knowledge,hernandez2023inspecting}. With much of work focused the goal of effectively editing information within LLMs  \cite{petroni-etal-2019-language,zhu2020modifying,Bau_2020,de-cao-etal-2021-editing,DBLP:journals/corr/abs-2305-13172}. In particular, one of the prominent recent trends has been to view the MLP weights as key-value memories which encode and recall facts \cite{geva2021transformer,geva2022transformer}. This analysis has been extended in \cite{meng2022locating,meng2022memit} to construct a knowledge location and editing scheme inspired by causal analysis in LLMs pioneered in \cite{vig2020causal}. The focus on multilayer-perceptrons (MLPs) has not been unique, other research has implicated attention as a factor in knowledge storage, while \cite{hase2023does} casts doubt on the efficacy of previous localization processes entirely. Curiously, this work does not appear to invalidate the editing methods, as they are confirmed to be effective. Nevertheless, others still question the usefulness of knowledge editing entirely \cite{youssef2023facts,brown2023edit,pinter2023emptying}. However, with the ability to edit knowledge by choosing areas that were not identified as being the most important, we return to the base idea of \textit{whether conceptual knowledge, rather than individual facts,} is encoded within related regions within these models.


\section{Conceptual Relatedness, Similarity, and Localization}\label{ProblemStatement}

\subsection{Conceptual Relatedness: Taxonomy and Meronomy}
\begin{figure}[tb]
  \centering
  \centerline{\includegraphics[height=5.0cm,width=8.5cm]{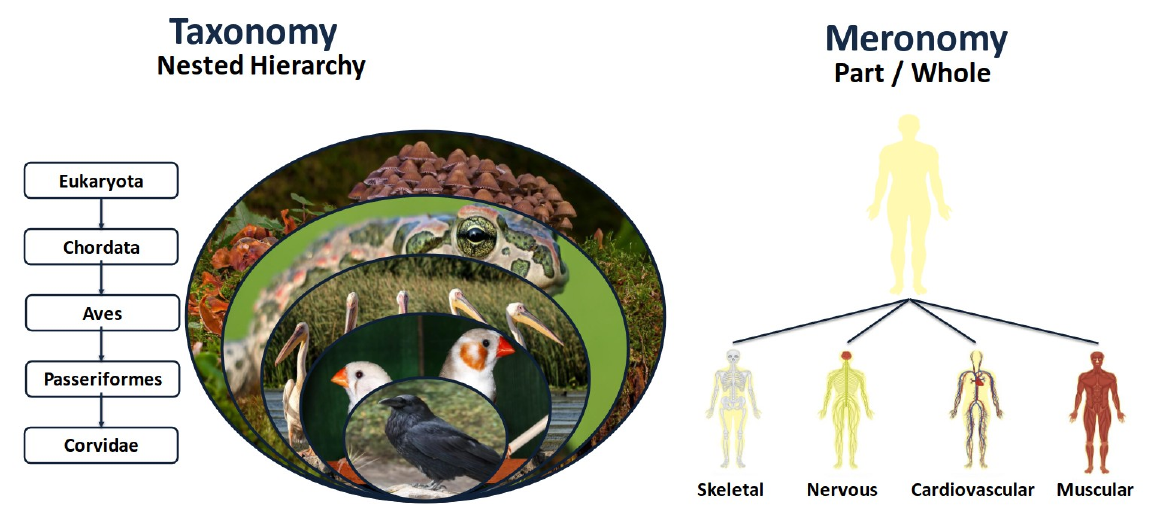}}

\caption{Hierarchical Types of Related Knowledge: Taxonomic and Meronomic Categorization} 
\label{fig:hierarchy}

\end{figure}

Our primary question is, \textit{``Does localization of concepts occur in GPT-like models?''} Here we define concept as a collection of related inputs, where the relation is a common association between the subject or object of the inputs. For example, consider the species the American crow (\textit{Corvus brachyrhynchos}) and the Fish crow (\textit{Corvus ossifragus}). Both species possess a fundamental, intuitive similarity between them, i.e both are crows. More rigorously, both species belong to the same genus \textit{Corvus}, where their inclusion has been determined by genetic similarity, among other physical attributes. These taxonomic relations (Left side of Figure \ref{fig:hierarchy})
of localization in models. Inputs within a level of their associated taxonomic tree 
should possess a correspondence with that of their indicated importance regions within a model, if such locality exists. 

\textit{Taxonomy} is therefore one of our choices for testing categorical localization. Our second choice is also a hierarchical relation, that of \textit{meronomy}, which is expressed as a part / whole relationship (Right side of Figure \ref{fig:hierarchy}). \textit{Meronomy} provides us an alternative expression of hierarchy in comparison to taxonomy as while all parts of a meronomic relationship are subsets of a greater whole, there is no ordering imposed on the subsets.

We note that hierarchical distinctions like taxonomy or meronomy just some of many types of relations that indicate similarity and that our choice of focusing on hierarchical categorization does not imply that such distinctions are not encoded within LLMs. We choose hierarchical relations due to their familiar nature, ease of use for generating quality data that can be assured to conform to the relation, and, especially for taxonomy, the innate ordering possessed by the expanding classifications.

%

\subsection{Similarity and Locality}

\subsubsection{Intra and Inter-Category Similarity.} We define our concept of similarity with respect to locality using the output of the causal analysis procedure (Section \ref{methodsForCausalAnalysis}) which calculates layer based importance
of an individual subject via the indirect effect. The \textit{deviation in the ordering of these layers is our base for measures of determining locality.} Regarding intra-category similarity, we evaluate the similarity among components within a category. For example, in the collection of living things we can compare the genus \textit{Corvus} (Crows, Ravens, etc.) and the order \textit{Passeriformes} (Birds in general). 
To prevent erroneously high levels of similarity, the data generated for any taxonomic level is mutually exclusive with all prior levels, i.e., the phylym \textit{Chordata} (Vertebrates and close relatives) excludes the prior level the class \textit{Aves} (Birds, which are vertebrates). 
In our meronomic data, this mutual exclusivity is inherent. Regarding inter-category similarity, we evaluate such similarity across category, using both individual layers within a category as well as aggregates of the entire category as items for comparison.

\subsubsection{Similarity and Concept Partition.} The idea of similarity is the foundation between the partitioning of related concepts. The whole purpose of classifiers is to make a judgment about similarity, a process which is inherent to human (and most other animals) cognition since pattern recognition is a necessary condition for basic survival. 
 And while neural networks in general were inspired by computation in biological systems, their process of pattern storage does \textit{not} need to conform solely to biological frameworks. For instance, the human brain can be divided into many sections, each with their own collection of important functions. With respect to pattern recognition, certain patterns are more strongly emphasized into particular regions, like the parietal lobe for object identification. While others may be distributed between regions, like with both the frontal and temporal lobes being indicated in smell recognition. Do LLMs possess similar sections? That is, are there regions of their constituent components more strongly associated with some action or concept? Just because components of our models were patterned on a representation of biological cognition does not imply there will be any such locality present. Some existing evidence  already show opposition against locality where layers of GPT-like models have been found to be interchangeable to some extent--i.e., the exact order of the layers is not particularly important for certain regions \cite{zhao2021nonlinearity}. \citealp{meng2022locating} reference this directly and additionally conjecture that \textit{any fact could equivalently be stored in any one of the middle MLP layers}.


\subsection{Methods for Causal Analysis} \label{methodsForCausalAnalysis}

To answer the question of concept localization, we appeal to causal methods developed for locating the important regions for a singular input. In particular our analysis is based directly on the works by Meng et al. whose goal was to locate important regions corresponding to individual facts to ultimately edit these facts effectively \cite{meng2022locating}. That work is focused on editing individual pieces of knowledge and so the collection of facts used do not require any particular theme. These facts are disparate,  which makes possible similarities between regions associated with related facts difficult to discern. Instead, we require a larger, dedicated dataset designed to have its constituent facts partitioned into distinct categories. We expand on the creation of this dataset next in Section \ref{sec:dataset}. As our analysis is directly based on methods developed by Meng et al. we refer readers to their paper for the causal analysis process \cite{meng2022locating}. 

\section{\dataset: A Dataset for Related Concepts}\label{sec:dataset}

As prior work in knowledge location has focused on the editing of individual facts, these facts tend to be fairly specific. Largely about an individual person or place, and often unrelated to each other, nor possessing any integral structures or themes. To begin to test the idea of categorical locality it is necessary to create a new collection of data, one deliberately segmented into clusters of closely related facts.

Based on the discussion in Section~\ref{ProblemStatement}, we focus the dataset on hierarchical relations, specifically taxonomy and meronomy. For each relation, we generate two datasets, with the goal of choosing specific concepts that have sufficient quantities of reliable data to generate thousands of facts while across each subcategory within the hierarchy. Each individual fact adheres to the format in \cite{meng2022locating} where a fact is comprised of three distinct parts. The subject $s$, the object $o$, and the relation $r$. With the subject of fact belonging directly to the associated category and the object being the correct result of the prediction of the model whose input is the subject plus the relation. 

All facts were generated using GPT 3.5 Turbo and GPT 4.0 (3.5 Turbo predominantly as initial testing showed no appreciable increase in fact quality between the versions). The fact generation process was broken into two parts. The first being the generation of raw facts associated with each subcategory within one of the four primary categories. This generation used triple-shot prompting of facts that conformed to the desired format and subcategory. The fact generation template, associated prompts and example facts can be found in Appendix \ref{apn:prompts}. 

\renewcommand{\tabcolsep}{1.5pt}
\begin{table}[t]
\caption*{\textbf{\dataset}: A Dataset for Related Concepts}
\centering
\footnotesize
\begin{tabular}{ccccc}
\toprule

\multicolumn{1}{c}{\textbf{Category}} & \multicolumn{1}{c}{\textbf{Hierarchical}} &
\multicolumn{1}{c}{\textbf{\# of}} & \multicolumn{1}{c}{\textbf{\# of}}& \multicolumn{1}{c}{\textbf{Predictive }}
\\
\multicolumn{1}{c}{\textbf{}} & \multicolumn{1}{c}{\textbf{Relation}} &
\multicolumn{1}{c}{\textbf{Concepts}} & \multicolumn{1}{c}{\textbf{Facts}}& \multicolumn{1}{c}{\textbf{ Accuracy}}
\\
\midrule
Birds & Taxonomic & 8 &  27157 &  23.0\%\\
Dogs & Taxonomic & 8 &  25274 & 20.0\%\\ 
Organ Systems & Meronomic & 11 &  39976 & 21.6\% \\
Auto. Systems & Meronomic & 9 &  36366 & 20.7\% \\
\bottomrule
\end{tabular}
\caption{Summary of \dataset. Predictive accuracy with respect to GPT-2-XL.}
\label{tab:dataset_summary}
\vspace{-10pt}
\end{table}

Queries were batched in collections of 100 and fed back into the next query to reduce duplicates and to reduce subject and format drift. No changes were made to parameters like temperature for the final data generation, as initial testing found that when deviated from the default setting, the LLMs produce too few unique facts, or erroneous or nonsensical results. The length of the facts was restricted to a soft maximum of 15 words, to both reduce the time needed to generate the data, as well as provide a soft limit on the complexity of the fact.
The second stage passes the generated facts back into GPT to select the subject, object, and relation between them then return a JSON formatted representation. To validate the quality of the generated facts, 25 facts were sampled form each subcategory and manually verified. Of the 850 verified facts only 3 were found to be excessively vague or otherwise inaccurate and were manually changed. No subcategory exceeds a threshold of 1\% error,  indicating that the quality of the facts generated is high.



                  
                  
                  
                  

\renewcommand{\tabcolsep}{1.5pt}
\begin{table*}[t]
\centering
\footnotesize
\begin{tabular}{lcc|lcc|lcc|lcc}
\toprule

\multicolumn{1}{c}{\textbf{Category}} & \multicolumn{2}{c}{\textbf{Max AIE Layer}} & \multicolumn{1}{c}{\textbf{Category}}&\multicolumn{2}{c}{\textbf{Max AIE Layer}} & \multicolumn{1}{c}{\textbf{Category}} & \multicolumn{2}{c}{\textbf{Max AIE Layer}}& \multicolumn{1}{c}{\textbf{Category}} & \multicolumn{2}{c}{\textbf{Max AIE Layer}}\\
\cmidrule(lr){1-12}
\multicolumn{1}{c}{\textbf{Birds}} &\multicolumn{1}{c}{\textbf{MLP}} & \textbf{ATN} & \multicolumn{1}{c}{\textbf{Dogs}} & \textbf{MLP}&\multicolumn{1}{c}{\textbf{ATN}} &\multicolumn{1}{c}{\textbf{Organs}} & \textbf{MLP} & \multicolumn{1}{c}{\textbf{ATN}} &\multicolumn{1}{c}{\textbf{Auto}} & \multicolumn{1}{c}{\textbf{MLP}} &\multicolumn{1}{c}{\textbf{ATN}}\\
\cmidrule(lr){2-2}\cmidrule(lr){5-5}\cmidrule(lr){3-3}\cmidrule(lr){6-6}\cmidrule(lr){9-9}\cmidrule(lr){8-8}\cmidrule(lr){11-11}\cmidrule(lr){12-12}

Am. Crow & 5 &  27 &   Dog & 16 &  30  &  Circulatory & 5 & 27 & Braking & 5 & 27 \\
Corvus & 4 &  30 &   Canis & 5 &  31  &  Digestive & 5 & 27 &  Drivetrain & 5 & 27  \\
Corvidae & 16 &30 &  Caniformia & 15 &  28  &  Endocrine & 5 & 27 & Electrical & 0 & 27   \\
Passeriformes & 5 & 30 &  Carnivora & 15 &  28  &  Integumentary & 5 & 30 & Engine & 5 & 27   \\
Aves & 15 & 30 & Mammalia & 15 &  28  &  Immune & 8 & 25 & Exhaust & 4 & 27   \\
Chordata* & 15  & 30 &   Chordata* & 5 &  27  &  Muscular & 1 & 27 & Fuel & 5 & 27  \\
Anamalia  & 5 & 30 &  Anamalia & 5 &  30 &  Nervous & 8 & 27 & Frame  & \multirow{2}{*}{5}  & \multirow{2}{*}{27}   \\
Eukaryota & 11 & 30 &  Eukaryota  & 11 &  30 &  Reproductive & 14  & 27  & \& Body & &    \\
& & & & & & Respiratory & 14  & 27  & Ignition  & 4  & 27   \\
& & & & & & Skeletal & 5  & 30  & Suspension & \multirow{2}{*}{5} & \multirow{2}{*}{27}  \\
& & & & & & Urinary & 5  & 27  & \& Steering &  &  \\
\textbf{Baseline} & 17 &  31  &  \textbf{Baseline} & 17 &  31  & \textbf{Baseline} & 17  & 31 & \textbf{Baseline} & 17  & 31
\\

\bottomrule
\end{tabular}
\caption{Layer of Maximum Average Indirect Effect - GPT-2-XL (*) The subcategory Chordata differs between category due to the requirement of mutual exclusivity between subcategories.}
\label{tab:max_layer_stats}
\vspace{-10pt}
\end{table*}

\subsection{Taxonomic Categories}
To satisfy both quantity and quality of available information, the taxonomic categories are based on the same structure, that of biological taxonomy. Each category uses the classic eight layer representation used to categorize life which follows in order of increasing specificity: \textit{Domain}, \textit{Kingdom}, \textit{Phylum}, \textit{Class}, \textit{Order}, \textit{Family}, \textit{Genus}, and \textit{Species}. Further differentiation, i.e subspecies or subphylum were omitted due to limitations of generating sufficient numbers of facts that were mutually exclusive with the previous layer.

The two categories (\textbf{dogs}, and \textbf{birds}) were based on two colloquial groupings with large amounts of viable facts and which possessed a sufficient intuitive difference between each. For dogs the base level (species) was the domestic dog and continues up through the class Mammalia. For birds the base level was the American crow and continues up through the class Aves (The largest taxa containing ``birds''). We note that the distinction between the two categories ends at the class level. Coarser levels above the phylum chordata (which contains the vertebrates in which both birds and dogs are a member of) are identical. The phylum chordata itself differs between categories due to fact generation excluding the previous class as viable subjects for facts. For example the subcategory chordata generated for birds here omits any member of the class aves, and so omits every more specific grouping of birds available (Figure \ref{fig:hierarchy}).

\subsection{Meronomic Categories}
Meronomic relations afford us more diversity in subject material as our requirement for mutual exclusion between subcategory facts is largely inherent. We again concentrate on a single theme, that of anatomy, but use two significantly different foundations, one biological (human) and one technological (automobiles). For human anatomy we focus on the eleven major divisions of human organ systems: \textit{Circulatory}, \textit{Digestive}, \textit{Endocrine}, \textit{Integumentary}, \textit{Immune}, \textit{Muscular}, \textit{Nervous}, \textit{Reproductive}, \textit{Respiratory}, \textit{Skeletal}, and \textit{Urinary}. For automobiles we use the following nine systems \textit{Braking}, \textit{Drive Train}, \textit{Electrical}, \textit{Engine}, \textit{Exhaust}, \textit{Fuel}, \textit{Frame and Body}, \textit{Ignition}, and \textit{Suspension and Steering}.
\newline

\noindent A general summary of \dataset $\;$ can be seen in Table \ref{tab:dataset_summary}. Extended information can be found in Appendix \ref{apn:dataset}. Note that the causal tracing process is only performed on successful predictions, the predictive accuracy within Table \ref{tab:dataset_summary} is with respect to GPT-2-XL and refers to the percentage of causally traced facts within each dataset. 

\section{Experiment Setup}
\subsection{Models}
The models used were three differently sized variants of GPT-2. Medium (345M parameters), Large (762M parameters), and Extra Large (1542M parameters). Larger models proved infeasible as the causal tracing process requires non-trivial amounts of computation for even moderate sized datasets like \dataset. GPT-2-Medium averaged approximately 2 hours per subcategory, GPT-2-Large 6 hours, and GPT-2-XL 16 hours. Models larger than GPT-2-XL proved untenable requiring in excess of a month of continuous computation (Using a single A6000) to completely trace \dataset. Our discussion concentrates on the largest model, GPT-2-XL, to provide the closest link with relevant related work which has been been centered on GPT-2-XL.
\subsection{Procedure}
The experimental procedure uses the same causal tracing process from \cite{meng2022locating}. Each category has its constituent subcategories causally traced with the restriction of only correct predictions being candidates for tracing. The output of the causal tracing concentrates on the indirect effect (IE) for each layer's components. Larger effects relative to other layers indicate more importance to be placed on that layer with respect to the (sub)category. The IE for each layer is further stratified based on the partition of the input to the location of the token(s) which are: The \textit{First Subject Token},\textit{ Middle Subject Token(s)}, \textit{Last Subject Token},\textit{ First Subsequent Token}, \textit{Further Tokens},\textit{ Last Token}. The IE is averaged over a batch of tracing to reduce variation due to the nosing process and so is generally reported as Average Indirect Effect (AIE). As prior work has determined that the strongest IE for a layer's MLP component is at the Last Subject Token we focus on the MLP tracing exclusively at this location. The attention components have also been indicated as locations for the storage of knowledge \cite{geva2023dissecting}, giving us another possibility to search for locality. Here we concentrate on the Last Token for Attention modification as this location has been shown to have the strongest IE.

\subsection{Evaluation Metrics}
The output of the causal tracing is presented in the following formats (The smaller GPT models provide similar results which can be found in Appendix \ref{apn:extra}). The maximally important layer can be seen in Table \ref{tab:max_layer_stats} (Extended versions in Appendix \ref{apn:extra}). Heat maps generated with respect to the AIE for each layer (Subsets seen in Figure \ref{fig:correlation_subset_examples}, Full versions in Appendix \ref{apn:extra}). Finally, line plots are generated using the unaltered process in prior work in order to provide a direct comparison of outputs. As it can be difficult to discern differences between the plotted lines with the numerous subcategories, we provide a non-parametric comparison through the use of Spearman correlation (Subsets seen in Figure \ref{fig:correlation_subset_examples}). The Spearman correlation returns the linear correlation of the order of the layers where a correlation of one denotes an identical ordering and negative one the complete inverse of the original. Additionally, we used the dataset from \cite{meng2022locating} and reran the tracing results, taking these as a baseline for comparison. This dataset is a collection of disparate facts ranging from geographical locations to notable individual persons or corporations.

\renewcommand{\tabcolsep}{3.5pt}
\begin{table}[]
\centering
\footnotesize
\begin{subtable}[t]{\textwidth}
\begin{tabular}{lcc|cc|cc|}
\toprule

\multicolumn{7}{c}{\textbf{Inter-Category Correlations}}   
\\
\midrule
\multicolumn{1}{c}{\textbf{Categories}} & \multicolumn{2}{c}{\textbf{GPT-2-M}} & \multicolumn{2}{c}{\textbf{GPT-2-L}}&\multicolumn{2}{c}{\textbf{GPT-2-XL}} 
\\



 &  \textbf{\underline{MLP}} & \textbf{\underline{ATN}} & \textbf{\underline{MLP}} & \textbf{\underline{ATN}} & \textbf{\underline{MLP}} & \textbf{\underline{ATN}} \cr  
 \addlinespace[3pt]
Birds \& Dogs & 0.91 & 0.79 & 0.76 & 0.66 & 0.79 & 0.72  \\
Birds \& Auto & 0.91 & 0.65 & 0.66 & 0.49 & 0.62 & 0.44   \\
Birds \& Organ & 0.89 & 0.67 & 0.74 & 0.69 & 0.55 & 0.68  \\
Organ \& Dogs & 0.88 & 0.74 & 0.70 & 0.73 & 0.58 & 0.73 \\
Organ \& Auto & 0.90 & 0.74 & 0.72 & 0.51 & 0.66 & 0.48   \\
Auto \& Dogs & 0.89 & 0.71 & 0.69 & 0.42 & 0.62 & 0.47 \\
Base \& Birds & 0.39 & 0.26 & 0.51 & 0.30 & 0.11 & 0.46  \\
Base \& Dogs & 0.45 & 0.44 & 0.34 & 0.46 & 0.11 & 0.47  \\
Base \& Auto & 0.46 & 0.68 & 0.60 & 0.56 & 0.36 & 0.49 \\
Base \& Organ & 0.46 & 0.56 & 0.55 & 0.55 & 0.18 & 0.68 \\

\bottomrule
\end{tabular}
\end{subtable}
\renewcommand{\tabcolsep}{3.25pt}

\begin{subtable}[t]{\textwidth}
\begin{tabular}{lcc|cc|cc|}

\multicolumn{7}{c}{\textbf{Intra-Category Correlations}}   
\\
\midrule
\multicolumn{1}{c}{\textbf{Categories}} & \multicolumn{2}{c}{\textbf{GPT-2-M}} & \multicolumn{2}{c}{\textbf{GPT-2-L}}&\multicolumn{2}{c}{\textbf{GPT-2-XL}} 
\\



 &  \textbf{\underline{MLP}} & \textbf{\underline{ATN}} & \textbf{\underline{MLP}} & \textbf{\underline{ATN}} & \textbf{\underline{MLP}} & \textbf{\underline{ATN}} \cr   
 \addlinespace[3pt]
Birds \& Birds & 0.92 & 0.79 & 0.51 & 0.61 & 0.82 & 0.66 \\
Dogs \& Dogs & 0.88 & 0.81 & 0.34 & 0.74 & 0.76 & 0.76 \\ 
Auto \& Auto & 0.90 & 0.75 & 0.60 & 0.81 & 0.66 & 0.69 \\ 
Organ \& Organ & 0.91 & 0.77 & 0.35 & 0.49 & 0.56 & 0.58 \\
\bottomrule
\addlinespace[-4pt]
\end{tabular}
\end{subtable}

\caption{Average inter-category and intra-category correlations}
\label{tab:category_correlations}
\vspace{-10pt}
\end{table}

\begin{figure*}[!htb]
\centering
\minipage{0.49\textwidth}
  \centerline{\includegraphics[width=1\linewidth,height=0.95\linewidth]{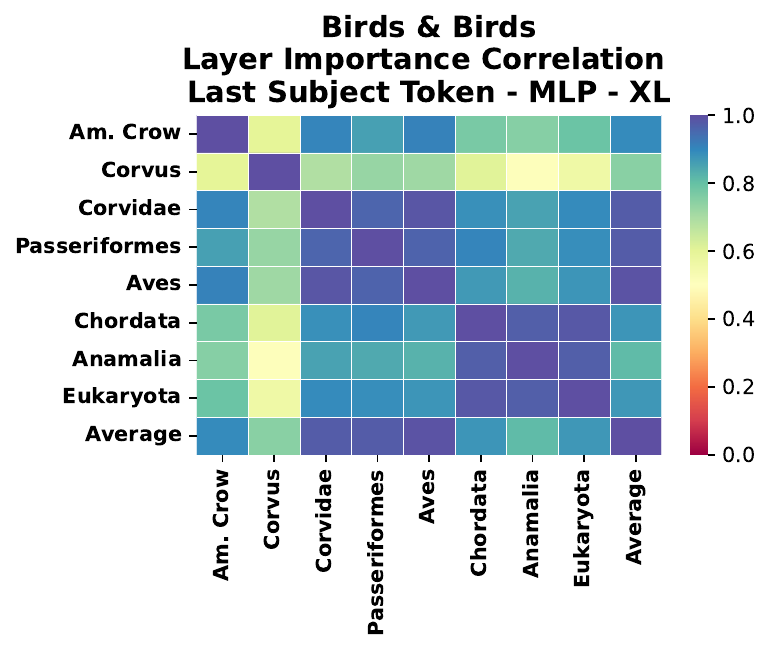}}
  \endminipage
\minipage{0.49\textwidth}  
  \centerline{\includegraphics[width=1\linewidth,height=0.95\linewidth]{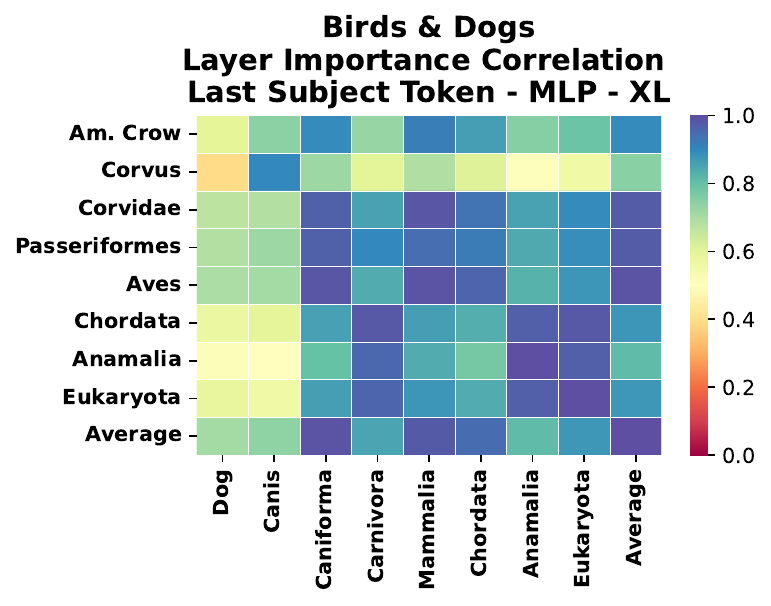}}
  \endminipage
\\
\minipage{0.49\textwidth}
  \centerline{\includegraphics[width=0.9\linewidth,height=0.9\linewidth]{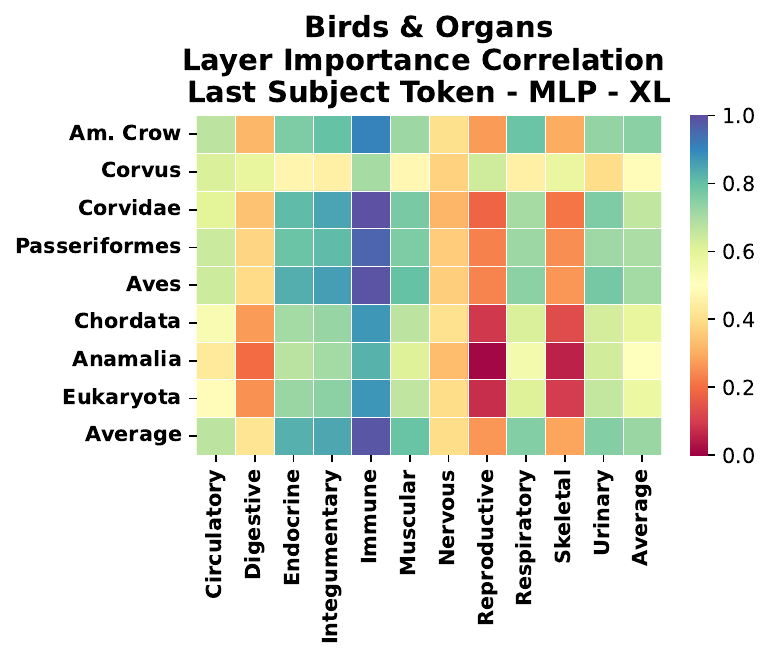}}
  \endminipage
\minipage{0.49\textwidth}  
  \centerline{\includegraphics[width=0.9\linewidth,height=0.90\linewidth]{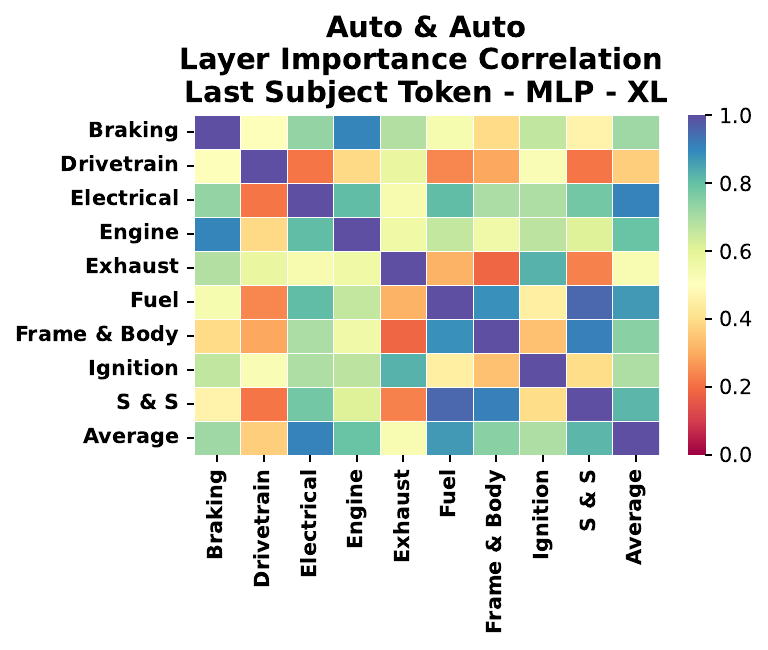}}
  \endminipage
\caption{Specific intra-category and inter-category correlation Heatmaps on GPT-2-XL} 
\label{fig:correlation_subset_examples}%
\vspace{-10pt}
\end{figure*}

\section{Results}
\subsection{Intra-Category Locality}
\noindent \textbf{Finding \#1: Taxonomic Locality:}
The taxonomic categories show some evidence of MLP specific intra-category locality. For GPT-2-XL the average intra-category correlations are 0.82 and 0.76 for Birds and Dogs, respectively (Table \ref{tab:category_correlations}). In comparison the (inter-category) correlations against the baseline are 0.11 and 0.11, respectively.

Visually, we see in the importance heatmap within category Birds overall high correlation and with clustering between related subcategories (Figure \ref{fig:correlation_subset_examples} Upper Left). Notably the three coarsest classifications of exclusively birds (\textit{Corvidae}, \textit{Passeriformes}, and \textit{Aves}) all show highly similar representations. The next three larger subcategories (\textit{Chordata}, \textit{Anamalia}, \textit{Eukaryota}) exhibit the same pattern. The most specific subcategories (\textit{American Crow} and \textit{Corvus}) are counter-intuitive, displaying appreciably less correlation between the other subcategories. Intra-category dogs exhibits remarkably close behavior to intra-category birds (Appendix \ref{apn:extra}). 

The meronomic categories show less evidence with average intra-category correlations of 0.66 and 0.56 for Organ and Auto systems, respectively.  The (inter-category) correlations against the baseline are 0.18 and 0.36, respectively. The heatmap within category Auto. Systems (Figure \ref{fig:correlation_subset_examples} Lower Right) shows less overall intra-category correlation in comparison to both taxonomic categories.  


\vspace{5pt}
\noindent \textbf{Finding \#2: Lack of Distinct Layer Importance for Individual Subcategories:}
When viewing the single layer of maximum importance for MLP interventions at the Last Subject Token we see no strong evidence of related regions for individual subcategories.  (Table \ref{tab:max_layer_stats}). That is, categories more similar to each other do not necessarily possess layers of maximal importance close to each other. We see there is no obvious pattern or order here among either the layer or the size of the effect. However, further analysis might indicate that the top-k layers possess some ordering, and the overall importance of the maximum layer here is not well understood. Additionally, the line plots generated possess extremely wide confidence intervals and the associated heat maps show no substantive patterns (Appendix \ref{apn:extra}). While there are clear patterns in terms of the ordering of the layers, the significance of minor changes to layer order remains in doubt due to prior work. For attention at the Last Token see no obvious patterns apart from a strong clustering towards the latter layers. The importance of later layers on attention is well known and is to be expected.


\begin{figure*}[tb]
\centering
\minipage{0.49\textwidth}  
  \centerline{\includegraphics[width=.80\linewidth,height=0.75\linewidth]{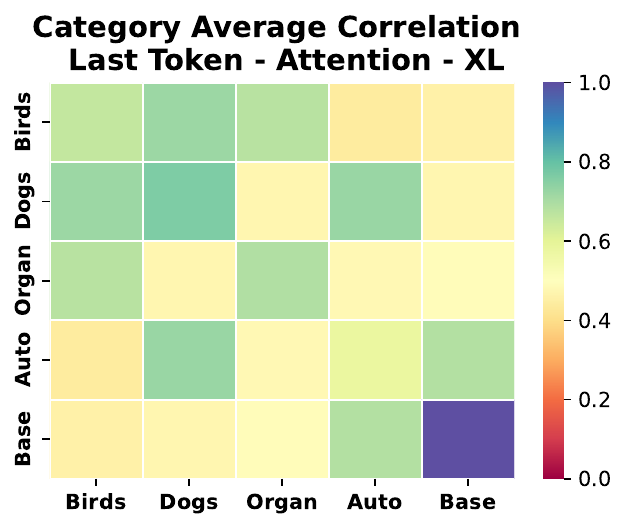}}
  \endminipage
\minipage{0.49\textwidth}  
  \centerline{\includegraphics[width=.80\linewidth,height=0.75\linewidth]{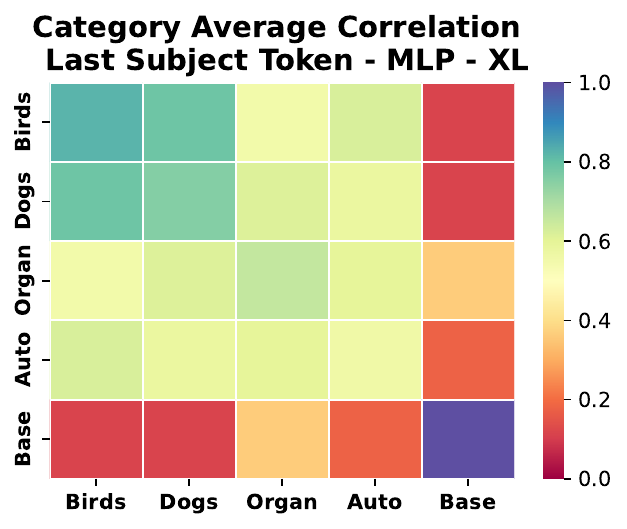}}
  \endminipage
\caption{Category average Spearman correlations on GPT-2-XL} 
\label{fig:category_average_correlation_example}%
\vspace{-10pt}
\end{figure*}

\subsection{Inter-Category Locality}

\noindent \textbf{Finding \#3: Taxonomic Locality:}

Similarly, the taxonomic categories also show initial signs of MLP focused inter-category locality between each other with Birds averaging a correlation of 0.79 between Dogs (Figure \ref{fig:correlation_subset_examples}, Table \ref{tab:category_correlations}) In comparison, the correlations between each of these categories with the baseline dataset is 0.11, substantially lower.
As these share two subcategories, \textit{Anamalia} and \textit{Eukaryota}, the previous values are computed without those subcategories. The taxonomic categories also show lesser correlation with the respective meronomic categories. Notably, \textbf{Auto} systems possesses very similar correlations between each of the taxonomic categories, and likewise with \textbf{Organ} systems (Figure \ref{fig:category_average_correlation_example}). Curiously, the baseline is the furthest from every other category which is intuitive as it is the least similar in nature to every other category. We note that attention focused locality possesses none of the above attributes and shows no distinctive patterning. Again our observations remain as above with the lack of substantive patterns within the layer of maximum importance.

\vspace{5pt}
\noindent \textbf{Finding \#4: Clustered Layer Importance:}
As with intra-category locality we see no patterns within the MLP interventions on the Last Subject Token indicative of progressively important layers. We do see tighter clustering, which is reasonable given the larger difference between the subcategories. Auto systems especially has noticeably small variation between layer max with every subcategory except for Electrical having max AIE at layers four or five. For attention interventions we see the same expected results as with intra-category locality.

\section{Discussion}
 While differences in layer orders are present with respect to locality, demonstrated by the differences in correlations, it is not known what amount, if any, makes the difference meaningful in terms of application. Additionally, the usual caveats regarding correlative measures apply, we have only demonstrated associations on a limited collection of models and data.

The baseline is notable for being distant from every aggregated category and nearly all individual subcategories. As the facts contained within the baseline do not comport to any of our relations this provides us with more evidence for inter-category locality. But the size and scope of the dataset are likely not sufficient enough to draw conclusions from. Curiously for the subcategories that correlate more closely with the baseline (\textit{Corvus}, \textit{Reproductive}, \textit{Skeletal}) these tend to correlate less with within their respective category. Why these particular subcategories do so is unknown, but increasing the size of these subcategories may rule out coincidence.

\section{Conclusion}
In the paper we have investigated knowledge locality, especially with regard to clusters of related facts. Overall, we have found evidence displaying both intra-category and inter-category locality for hierarchical data, particularly for taxonomic relations. However, we do not see evidence for an ordered representation based on layer importance that will allow targeting of specific clusters of information. What locality we find is broad and manifests as a strong correlation between the important layers across the entirety of the model.

\section{Limitations}

\begin{enumerate}[leftmargin=\dimexpr\parindent-0.2\labelwidth\relax,itemsep=2pt]
\item \textit{Transferability to other GPT-like models}: Due to the computational constraints listed above, the analysis was restricted to only three versions of GPT-2, Medium, Large, and XL. How the existence of category locality or its properties differs other architectures is unknown.


\item \textit{The importance of indirect effect size}: The magnitude of the average indirect effect tends to be low with values generally under 10\%. As the AIE is the fundamental measure used here to determine layer importance it is unknown if these values are too low to draw conclusions from. Baseline AIE values however tend to be lower than the \dataset $\;$ subcategories. 

\item \textit{Other methods of similarity}: We note that hierarchical distinctions like taxonomy or meronomy just some of many types of relations that indicate similarity and that our choice of focusing on hierarchical categorization does not imply that such distinctions are not encoded within LLMs. We choose hierarchical relations due to their familiar nature, ease of use for generating quality data that can be assured to conform to the relation, and, especially for taxonomy, the innate ordering possessed by the expanding classifications.
\end{enumerate}

\section*{Broader Impacts and Ethics Statement}
The authors anticipate there to be no reasonable
cause to believe use of this work would result in
harm, both direct or implicit. The authors disclaim
any conflicts of interest pertaining to this work.


\bibliography{anthology,references}
\newpage
\onecolumn

\appendix

\section{DARC - Additional Information} \label{apn:dataset}
\renewcommand{\tabcolsep}{8pt}
\begin{table*}[h]
\centering
\footnotesize
\begin{tabular}{lccc|lccc}
\toprule

\multicolumn{1}{c}{\textbf{}} &
\multicolumn{5}{c}{\textbf{Correct Predictions}} \\
\midrule

\multicolumn{1}{c}{\textbf{Category}} &\multicolumn{1}{c}{\textbf{GPT-2-M}} & \textbf{GPT-2-L} & \multicolumn{1}{c}{\textbf{GPT-2-XL}} &\multicolumn{1}{c}{\textbf{Category}} &\multicolumn{1}{c}{\textbf{GPT-2-M}} & \textbf{GPT-2-L} & \multicolumn{1}{c}{\textbf{GPT-2-2-XL}} \\
\cmidrule(lr){2-2}\cmidrule(lr){5-5}\cmidrule(lr){3-3}\cmidrule(lr){4-4}\cmidrule(lr){7-7}\cmidrule(lr){8-8}\cmidrule(lr){6-6}


Am. Crow & 417 & 445 & 475   &  Dog & 299 & 338 & 362   \\
Corvus & 700 &  768 & 830  &  Canis & 718 & 750 & 836    \\
Corvidae & 687 & 748 & 848  & Caniformia & 365 & 322 & 393   \\
Passeriformes & 616 & 686 & 767   & Carnivora & 557 & 482 & 549   \\
Aves & 514 & 610  & 706  & Mammalia & 652 & 418 & 849    \\
Chordata* & 816 & 889 & 1007 & Chordata* & 453 & 421 & 474   \\
Anamalia  & 725 & 795 &  903 &  Anamalia & 725 & 795 &  903   \\
Eukaryota & 559 & 604 &  648 & Eukaryota  & 559 & 604 & 648   \\
\textbf{Baseline} & 617 & 695 & 1208  &  \textbf{Baseline} & 617 & 695 & 1208 
\\

\bottomrule
\end{tabular}
\caption{Correct Predictions - Taxonomy}
\label{tab:DARC_Stats_1}
\vspace{-20pt}
\end{table*}

\renewcommand{\tabcolsep}{8pt}
\begin{table*}[h]
\centering
\footnotesize
\begin{tabular}{lccc|lccc}
\toprule

\multicolumn{1}{c}{\textbf{}} &
\multicolumn{5}{c}{\textbf{Correct Predictions}} \\
\midrule

\multicolumn{1}{c}{\textbf{Category}} &\multicolumn{1}{c}{\textbf{GPT-2-M}} & \textbf{GPT-2-L} & \multicolumn{1}{c}{\textbf{GPT-2-XL}} &\multicolumn{1}{c}{\textbf{Category}} &\multicolumn{1}{c}{\textbf{GPT-2-M}} & \textbf{GPT-2-L} & \multicolumn{1}{c}{\textbf{GPT-2-XL}} \\
\cmidrule(lr){2-2}\cmidrule(lr){5-5}\cmidrule(lr){3-3}\cmidrule(lr){4-4}\cmidrule(lr){7-7}\cmidrule(lr){8-8}\cmidrule(lr){6-6}

Circulatory & 749 & 815 & 847   & Braking & 800 & 904  &  958\\
Digestive & 832 &  902 & 756  &  Drivetrain & 812 & 865  &  914 \\
Endocrine & 560 & 563 & 474  & Electrical & 945 & 1012  &  1001 \\
Integumentary & 394 & 429 & 464 & Engine & 790 & 858  &  892 \\
Immune & 574 & 620  & 798   & Exhaust & 956 & 968  &  1001 \\
Muscular & 679 & 761 & 836  & Fuel & 542 & 608  &  622 \\
Nervous & 563 & 600 & 841 & Frame  & \multirow{2}{*}{503} & \multirow{2}{*}{541} & \multirow{2}{*}{644}    \\
Reproductive & 931 & 1020 & 1099 & \& Body & &  &     \\
Respiratory & 950 & 987 & 1060  & Ignition  & 702 & 766  &  766  \\
Skeletal & 724 & 791 & 899  & Suspension & \multirow{2}{*}{651} & \multirow{2}{*}{669} & \multirow{2}{*}{754}  \\
Urinary & 448 & 485 & 544  & \& Steering &  & &    \\
\textbf{Baseline} & 617 & 695 & 1208    &  \textbf{Baseline} & 617 & 695 & 1208    \\

\bottomrule
\end{tabular}
\caption{Correct Predictions - Meronomy}
\label{tab:DARC_Stats_2}
\vspace{-20pt}
\end{table*}

\renewcommand{\tabcolsep}{4pt}
\begin{table*}[hbt]
\centering
\footnotesize
\begin{tabular}{lclclclc}
\toprule


\multicolumn{2}{c}{\textbf{Birds}} & \multicolumn{2}{c}{\textbf{Dogs}} &
\multicolumn{2}{c}{\textbf{Organ Systems}} & \multicolumn{2}{c}{\textbf{Auto. Systems}}
\\

\cmidrule(lr){1-2}\cmidrule(lr){3-4}\cmidrule(lr){5-6}\cmidrule(lr){7-8}
\multicolumn{1}{c}{\textbf{Subcategory}} & \textbf{Total Facts} & \multicolumn{1}{c}{\textbf{Subcategory}} & \textbf{Total Facts}& \multicolumn{1}{c}{\textbf{Subcategory}} & \textbf{Total Facts} & \multicolumn{1}{c}{\textbf{Subcategory}} & \textbf{Total Facts} \\
\cmidrule(lr){1-1}\cmidrule(lr){2-2}\cmidrule(lr){3-3}\cmidrule(lr){4-4}\cmidrule(lr){5-5}\cmidrule(lr){6-6}\cmidrule(lr){7-7}\cmidrule(lr){8-8}
American Crow & 3441 & Domestic Dog & 3301 & Circulatory & 3813 & Braking & 3406 \\
Corvus & 3520 & Canis & 4018   & Digestive & 3436 & Drivetrain & 3836   \\
Corvidae & 3607 & Caniformia & 2510  & Endocrine & 2717 & Electrical & 4277  \\
Passeriformes & 3888 & Carnivora & 3252 & Integumentary & 3200 & Engine & 4159  \\
Aves & 3103  & Mammalia & 3083 & Immune & 3585  & Exhaust & 4876  \\
Chordata* & 3207 & Chordata* & 2728  & Muscular & 3638 & Fuel & 4048  \\
Anamalia & 3207 & Anamalia & 3207  & Nervous & 3931 & Frame  & \multirow{2}{*}{4222}   \\
Eukaryota & 3148 & Eukaryota & 3148  & Reproductive & 4254 & \& Body &  \\
 & & & & Respiratory & 4433 &  Ignition  & 4004  \\
 & & & & Skeletal & 3821 &  Suspension & \multirow{2}{*}{3538}  \\
 & & & & Urinary & 3148 & \& Steering &   \\ \\
 \textbf{Category Total:} & \textbf{27157} &  & \textbf{25274} & & \textbf{39976} & & \textbf{36366} \\

\bottomrule
\end{tabular}
\caption{Dataset Statistics for Taxonomic and Meronomic Categories. (*) The subcategory Chordata differs between category due to the requirement of mutual exclusivity between subcategories.}
\label{tab:DARC_Taxonomy}
\vspace{-10pt}
\end{table*}

\clearpage
\section{Fact Generation Prompt Template (Respiratory System)}\label{apn:prompts}
\subsection{Initial Generation}
\begin{verbatim}
    
Generate X facts about the human body's respiratory system.

Limit the size of each fact to no more than 15 words.
Do not include punctuation in any fact.
Do not reuse any of the previous facts provided.

Here is an example of three facts and their correct formatting:

The first bronchi to branch from the trachea are the right and left main bronchi

The lungs expand and contract during the breathing cycle drawing air in and out of the

  lungs

Gas exchange in the lungs occurs in millions of small air sacs
\end{verbatim}
\newpage
\subsection{Subject Identification and JSON Conversion}
\begin{verbatim}
    Interpret and then format each of the provided facts about
    X according to the following json format:
        
        {
           "known_id": NONE',
           "full_fact:": "COMPLETE FACT",
           "subject": "SUBJECT OF FACT",
           "attribute": "OBJECT OF FACT or FINAL WORD OF FACT",
           "prediction": "OBJECT OF FACT or FINAL WORD OF FACT",
           "prompt": "ALL OF full_fact UP TO BUT NOT INCLUDING THE
            OBJECT OF FACT OR THE FINAL WORD OF FACT",
           "group": "CATEGORY OF SUBJECT",
           "relation_id": "NONE",
           "template": "NONE",
           
         },
        
        
    Here are three examples of facts with their correct JSON formatting:
   
    The first bronchi to branch from the trachea are the right and left main bronchi

         {
            "known_id": NONE,
            "full_fact": "The first bronchi to branch from the trachea
             are the right and left main bronchi",
            "subject": "bronchi",
            "attribute": "bronchi",
            "prediction": "bronchi",
            "prompt": "The first bronchi to branch from the trachea
             are the right and left main",
            "group": "Respiratory",
            "template": "NONE",
            "relation_id": "NONE",
            
          },

        ...
        
        The facts to be coverted are: ...
            
\end{verbatim}

\newpage
\subsection{Example Facts - Respiratory System}

\begin{verbatim}
    
"full_fact": "Pneumothorax is a condition where air collects in the pleural space".

"full_fact": "Breathing efficiency varies with physical fitness",
 
"full_fact": "Carbon monoxide poisoning prevents oxygen transport leading to tissue
  hypoxia",

"full_fact": "Inhalation is an active process driven by muscle contractions",

"full_fact": "The nasal cavity helps warm and humidify inhaled air before it reaches
  the lungs",

"full_fact": "Hypoxia results from insufficient oxygen", 

"full_fact": "Pulmonary surfactant reduces surface tension in the alveoli",

"full_fact": "Respiratory alkalosis can occur due to rapid breathing causing a rise in 
  blood pH",

"full_fact": "The pleura create a friction-reducing surface for lung expansion within
  the chest cavity",

"full_fact": "Auscultation of the lungs can detect abnormal breath sounds indicating 
  respiratory issues",

"full_fact": "Bronchioles are smaller branches within the lungs leading to the alveoli",

"full_fact": "The phrenic nerve controls the diaphragm's movement for breathing",

"full_fact": "The trachea is lined with cilia to help sweep mucus and foreign 
  particles",
  
"full_fact": "The alveolar sacs are clustered at the end of the bronchioles for gas 
  exchange",

"full_fact": "Paranasal sinuses help humidify and warm air before it reaches the 
  lungs",
    
\end{verbatim}

\clearpage
\section{Additional Experimental Data} \label{apn:extra}

\subsection{GPT-2-XL}

\renewcommand{\tabcolsep}{0.5pt}
\begin{table*}[h]
\centering
\footnotesize
\begin{tabular}{lcc|cc|lcc|cc}
\toprule

\multicolumn{1}{c}{\textbf{Category}} & \multicolumn{2}{c}{\textbf{Last Subject Token (MLP)}} & \multicolumn{2}{c}{\textbf{Last Token (ATN)}}&\multicolumn{1}{c}{\textbf{Category}} & \multicolumn{2}{c}{\textbf{Last Subject Token (MLP)}} & \multicolumn{2}{c}{\textbf{Last Token (ATN)}}\\

\cmidrule(lr){1-1}\cmidrule(lr){2-3}\cmidrule(lr){4-5}\cmidrule(lr){6-6}\cmidrule(lr){7-8}\cmidrule(lr){9-10}
\multicolumn{1}{c}{\textbf{}} &\multicolumn{1}{c}{\textbf{Max at Layer}} & \textbf{A.I.E} & \multicolumn{1}{c}{\textbf{Max at Layer}} & \textbf{A.I.E}&\multicolumn{1}{c}{\textbf{}} &\multicolumn{1}{c}{\textbf{Max at Layer}} & \textbf{A.I.E} & \multicolumn{1}{c}{\textbf{Max at Layer}} & \textbf{A.I.E} \\
\cmidrule(lr){2-2}\cmidrule(lr){5-5}\cmidrule(lr){3-3}\cmidrule(lr){4-4}\cmidrule(lr){7-7}\cmidrule(lr){8-8}\cmidrule(lr){9-9}\cmidrule(lr){10-10}

Am. Crow & 5 & 0.142 & 27  & 0.087 &  Dog & 16 & 0.140 & 30  & 0.149 \\
Corvus & 4 &  0.079 & 30 & 0.074 &  Canis & 5 & 0.077 & 31  & 0.094   \\
Corvidae & 16 & 0.052 & 30 & 0.058 & Caniformia & 15 & 0.125 & 28  & 0.124  \\
Passeriformes & 5 & 0.049 & 30 & 0.057 & Carnivora & 15 & 0.054 & 28  & 0.057  \\
Aves & 15 & 0.082  & 30 & 0.086 & Mammalia & 15 & 0.111 & 28  & 0.093  \\
Chordata* & 15 & 0.070 & 30 & 0.071 & Chordata* & 5 & 0.130 & 27  & 0.117  \\
Anamalia  & 5 & 0.076 & 30 & 0.069 &  Anamalia & 5 & 0.076 & 30 & 0.069  \\
Eukaryota & 11 & 0.054 & 30 & 0.052 & Eukaryota  & 11 & 0.054 & 30 & 0.052  \\
\textbf{Baseline} & 17 & 0.056 & 31  & 0.086 &  \textbf{Baseline} & 17 & 0.056 & 31  & 0.086
\\

\bottomrule
\end{tabular}
\caption{Taxonomic Statistics - GPT-2-XL}
\label{tab:taxonomicGPT2XL}
\vspace{-10pt}
\end{table*}

\renewcommand{\tabcolsep}{0.5pt}
\begin{table*}[h]
\centering
\footnotesize
\begin{tabular}{lcc|cc|lcc|cc}
\toprule

\multicolumn{1}{c}{\textbf{Category}} & \multicolumn{2}{c}{\textbf{Last Subject Token (MLP)}} & \multicolumn{2}{c}{\textbf{Last Token (ATN)}}&\multicolumn{1}{c}{\textbf{Category}} & \multicolumn{2}{c}{\textbf{Last Subject Token (MLP)}} & \multicolumn{2}{c}{\textbf{Last Token (ATN)}}\\

\cmidrule(lr){1-1}\cmidrule(lr){2-3}\cmidrule(lr){4-5}\cmidrule(lr){6-6}\cmidrule(lr){7-8}\cmidrule(lr){9-10}
\multicolumn{1}{c}{\textbf{}} &\multicolumn{1}{c}{\textbf{Max at Layer}} & \textbf{A.I.E} & \multicolumn{1}{c}{\textbf{Max at Layer}} & \textbf{A.I.E}&\multicolumn{1}{c}{\textbf{}} &\multicolumn{1}{c}{\textbf{Max at Layer}} & \textbf{A.I.E} & \multicolumn{1}{c}{\textbf{Max at Layer}} & \textbf{A.I.E} \\
\cmidrule(lr){2-2}\cmidrule(lr){5-5}\cmidrule(lr){3-3}\cmidrule(lr){4-4}\cmidrule(lr){7-7}\cmidrule(lr){8-8}\cmidrule(lr){9-9}\cmidrule(lr){10-10}

Circulatory & 5 & 0.067 & 27  & 0.109 & Braking & 5 & 0.051 & 27  & 0.073 \\
Digestive & 5 &  0.080 & 27 & 0.097 &  Drivetrain & 5 & 0.053 & 27  & 0.077  \\
Endocrine & 5 & 0.079 & 27 & 0.104 & Electrical & 0 & 0.048 & 27  & 0.083  \\
Integumentary & 5 & 0.066 & 30 & 0.082 & Engine & 5 & 0.083 & 27  & 0.109  \\
Immune & 8 & 0.051  & 25 & 0.067 & Exhaust & 4 & 0.048 & 27  & 0.069  \\
Muscular & 1 & 0.092 & 27 & 0.107 & Fuel & 5 & 0.074 & 27  & 0.073 \\
Nervous & 8 & 0.037 & 27 & 0.061 & Frame  & \multirow{2}{*}{5} & \multirow{2}{*}{0.031} & \multirow{2}{*}{27}  & \multirow{2}{*}{0.027}   \\
Reproductive & 14 & 0.059 & 27 & 0.099 & \& Body & &  &   &  \\
Respiratory & 14 & 0.051 & 27 & 0.100 & Ignition  & 4 & 0.062 & 27  & 0.086   \\
Skeletal & 5 & 0.075 & 30 & 0.106 & Suspension & \multirow{2}{*}{5} & \multirow{2}{*}{0.050} & \multirow{2}{*}{27}  & \multirow{2}{*}{0.066}  \\
Urinary & 5 & 0.094 & 27 & 0.113 & \& Steering &  & &  &  \\
\textbf{Baseline} & 17 & 0.056 & 31  & 0.086 &  \textbf{Baseline} & 17 & 0.056 & 31  & 0.086 \\
\bottomrule
\end{tabular}
\caption{Meronomic Statistics - GPT-2-XL}
\label{tab:meronomic_statsGPT2XL}
\vspace{-10pt}
\end{table*}

\begin{figure*}[h]
  \centering
  \centerline{\includegraphics[height=8.25cm,width=\linewidth]{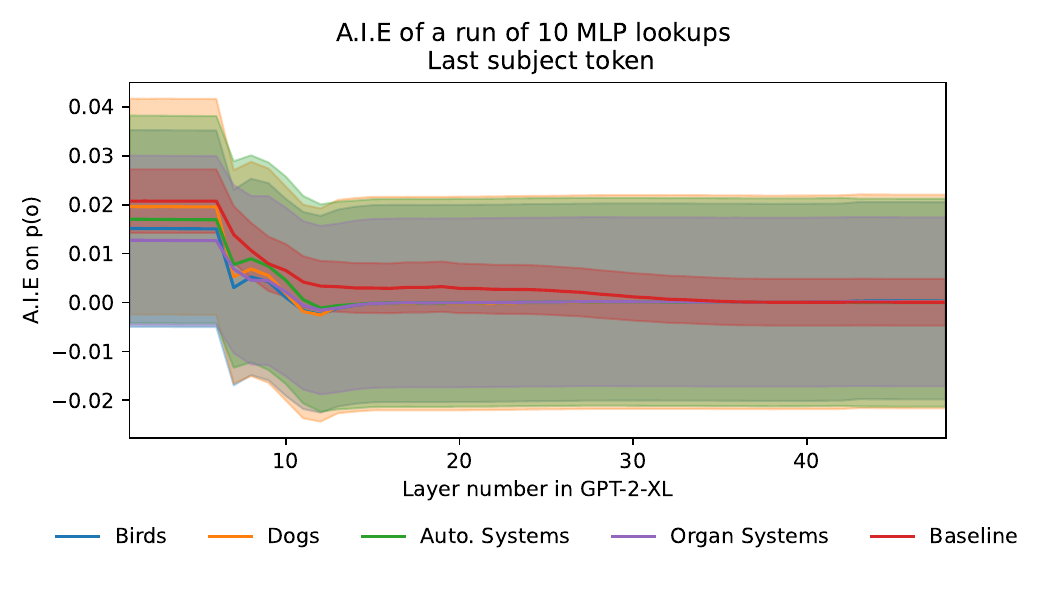}}

\caption{GPT-2-XL A.I.E Category Average - Last Token - MLP}
\label{fig:cat_avg_xl_mlp}
\end{figure*}

\begin{figure*}[h]
  \centering
  \centerline{\includegraphics[height=8.25cm,width=\linewidth]{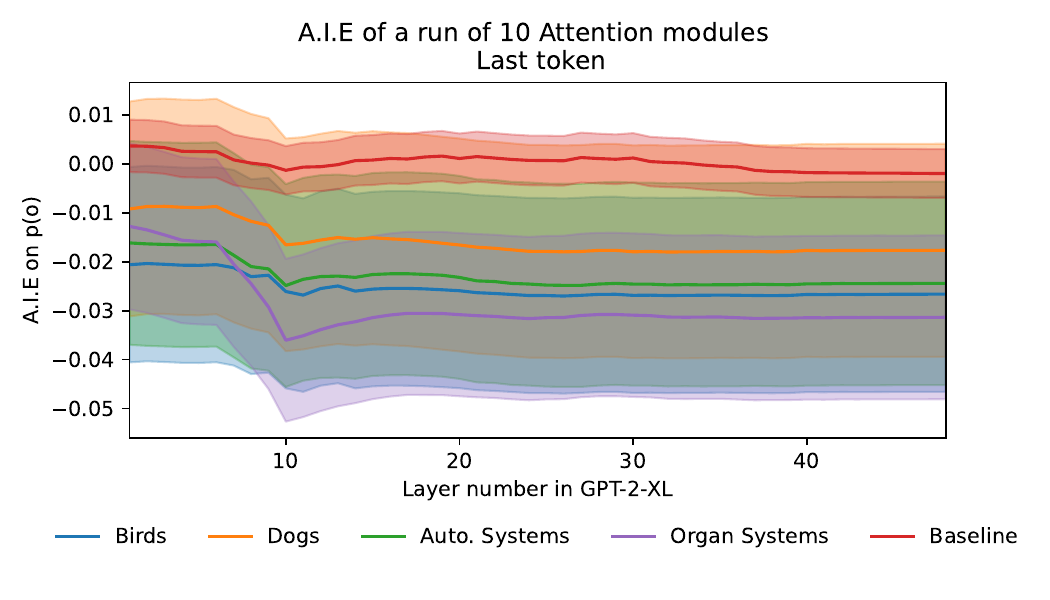}}

\caption{GPT-2-XL A.I.E Category Average - Last Token - Attention}
\label{fig:cat_avg_xl_attn}
\end{figure*}

\begin{figure*}[]
  \centering
  \centerline{\includegraphics[width=\linewidth]{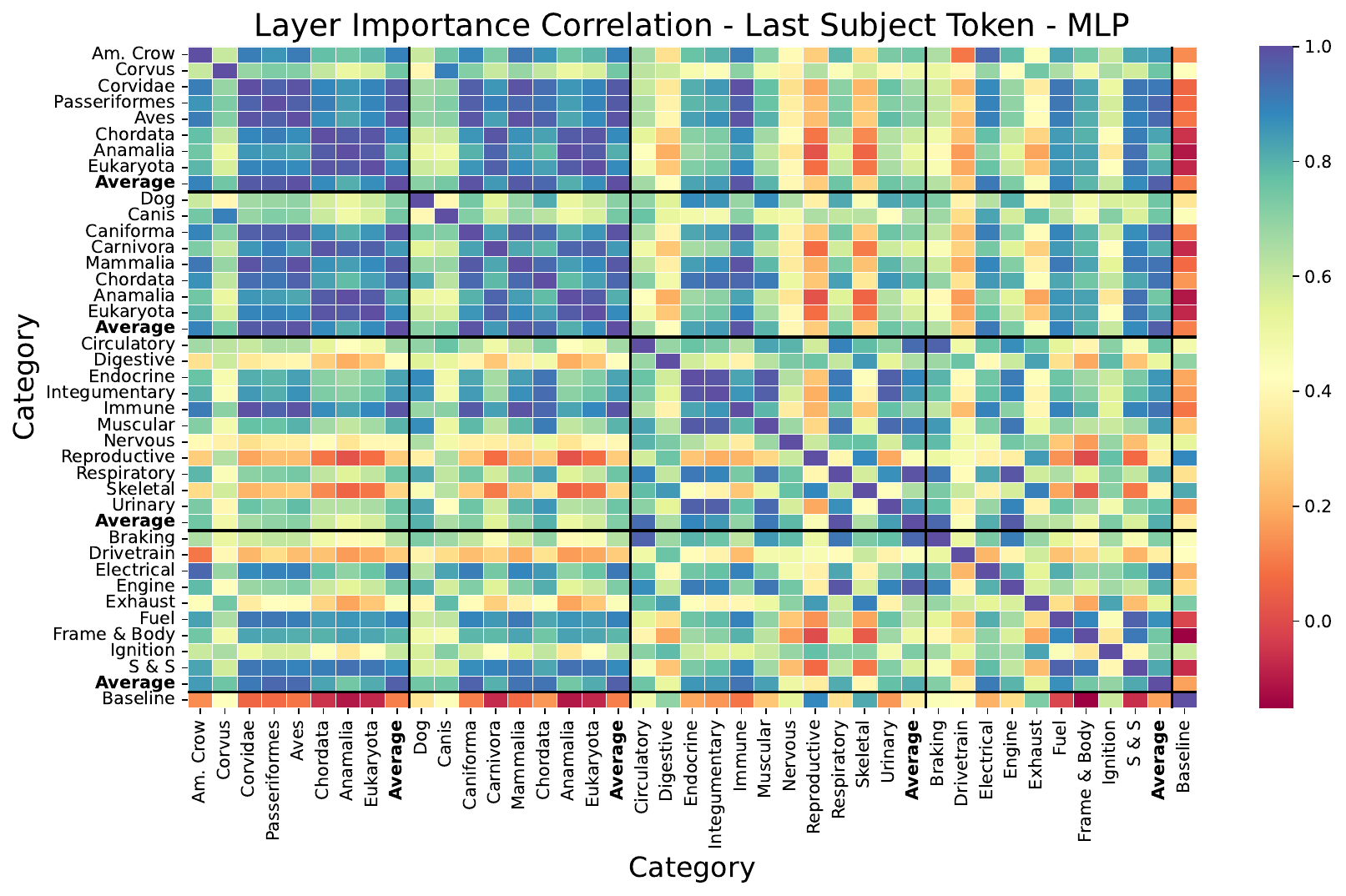}}

\caption{GPT-2-XL Layer Importance Heat Map - Last Subject Token - MLP}
\label{fig:LIH-XL-MLP}

\end{figure*}

\begin{figure*}[]
  \centering
  \centerline{\includegraphics[width=\linewidth]{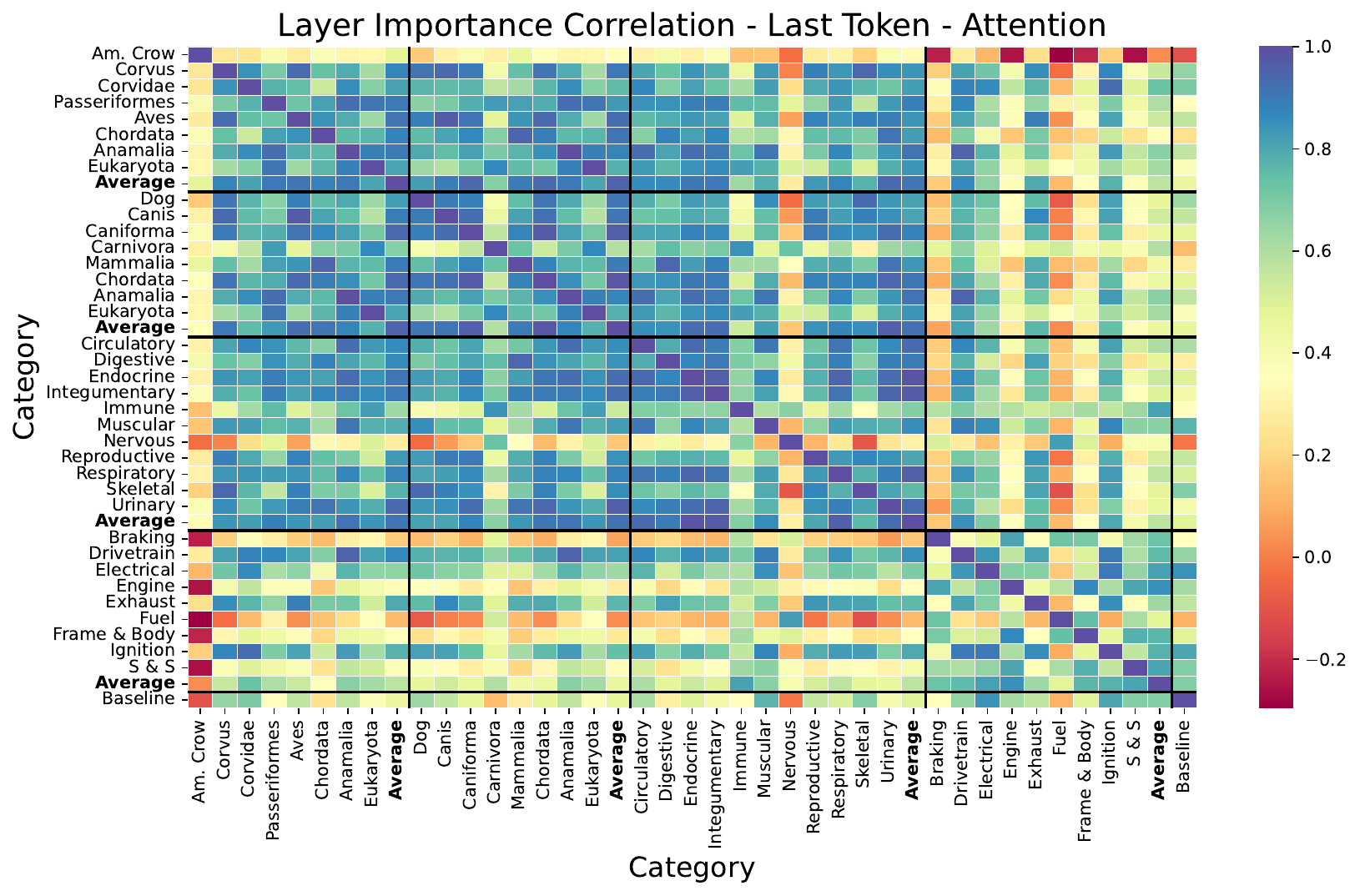}}

\caption{GPT-2-XL Layer Importance Heat Map - Last Token - Attention}
\label{fig:LIH-XL-Attn}

\end{figure*}

\begin{figure}[b]
    \centering
    \begin{minipage}{0.48\textwidth}
        \centering
        \includegraphics[width=1.0\textwidth]{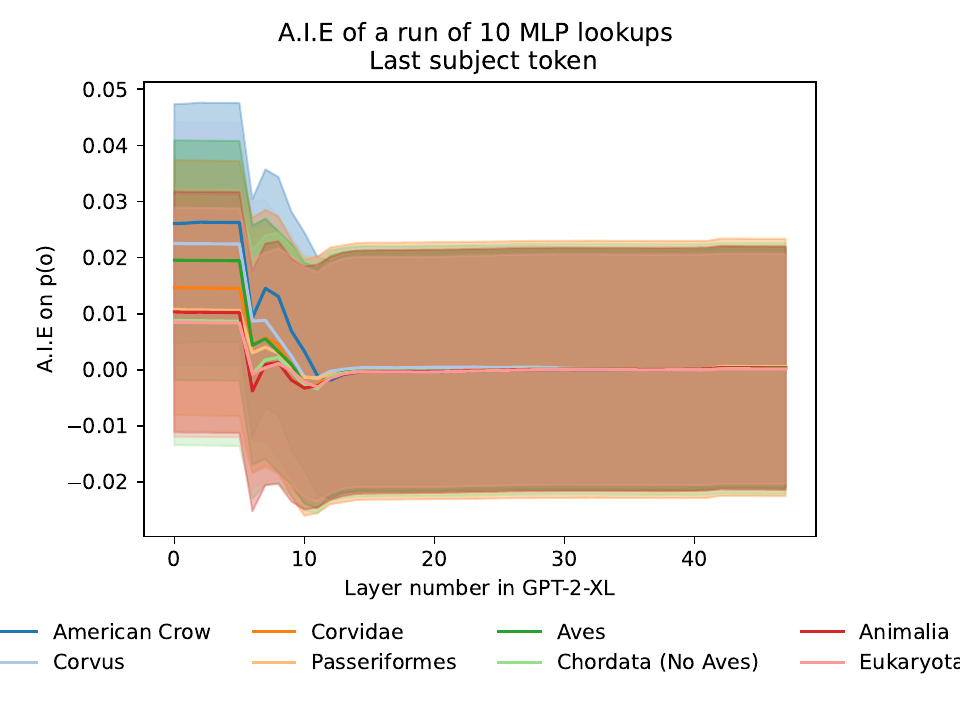} 
    \end{minipage}\hfill
    \begin{minipage}{0.48\textwidth}
        \centering
        \includegraphics[width=1.0\textwidth]{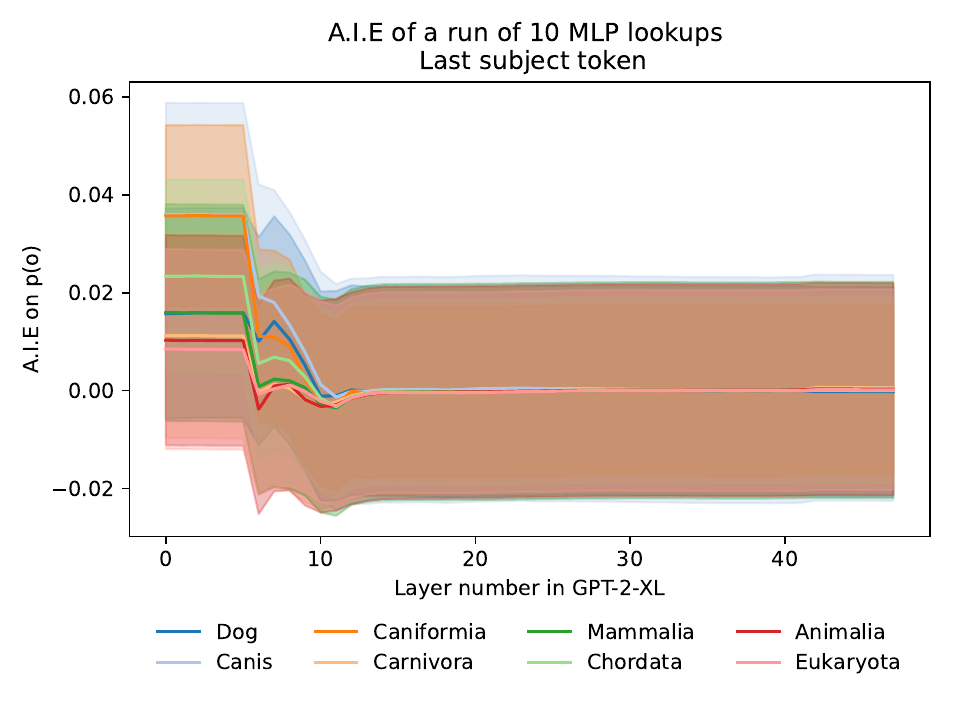} 
    \end{minipage}
    \begin{minipage}{0.48\textwidth}
        \centering
        \includegraphics[width=1.0\textwidth]{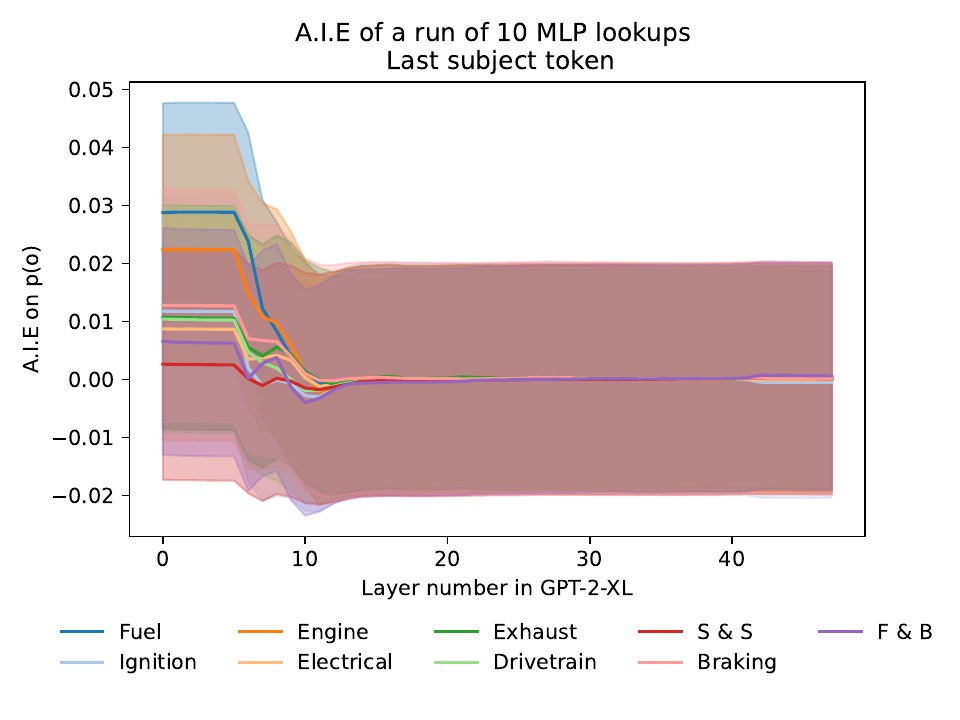} 
    \end{minipage}\hfill
    \begin{minipage}{0.48\textwidth}
        \centering
        \includegraphics[width=1.0\textwidth]{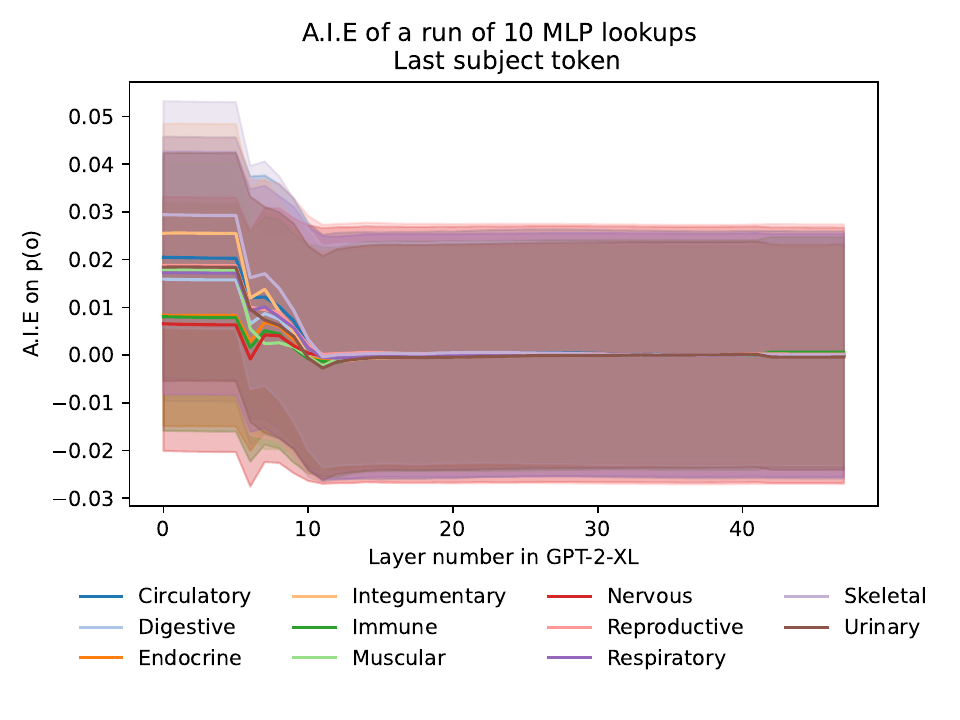} 
    \end{minipage}

\caption{GPT-2-XL Line Plots - Last Subject Token - MLP }
\label{fig:GPT2XL_Lineplots_MLP}
\end{figure}

\begin{figure*}[b]
  \centering
  \centerline{\includegraphics[width=\linewidth]{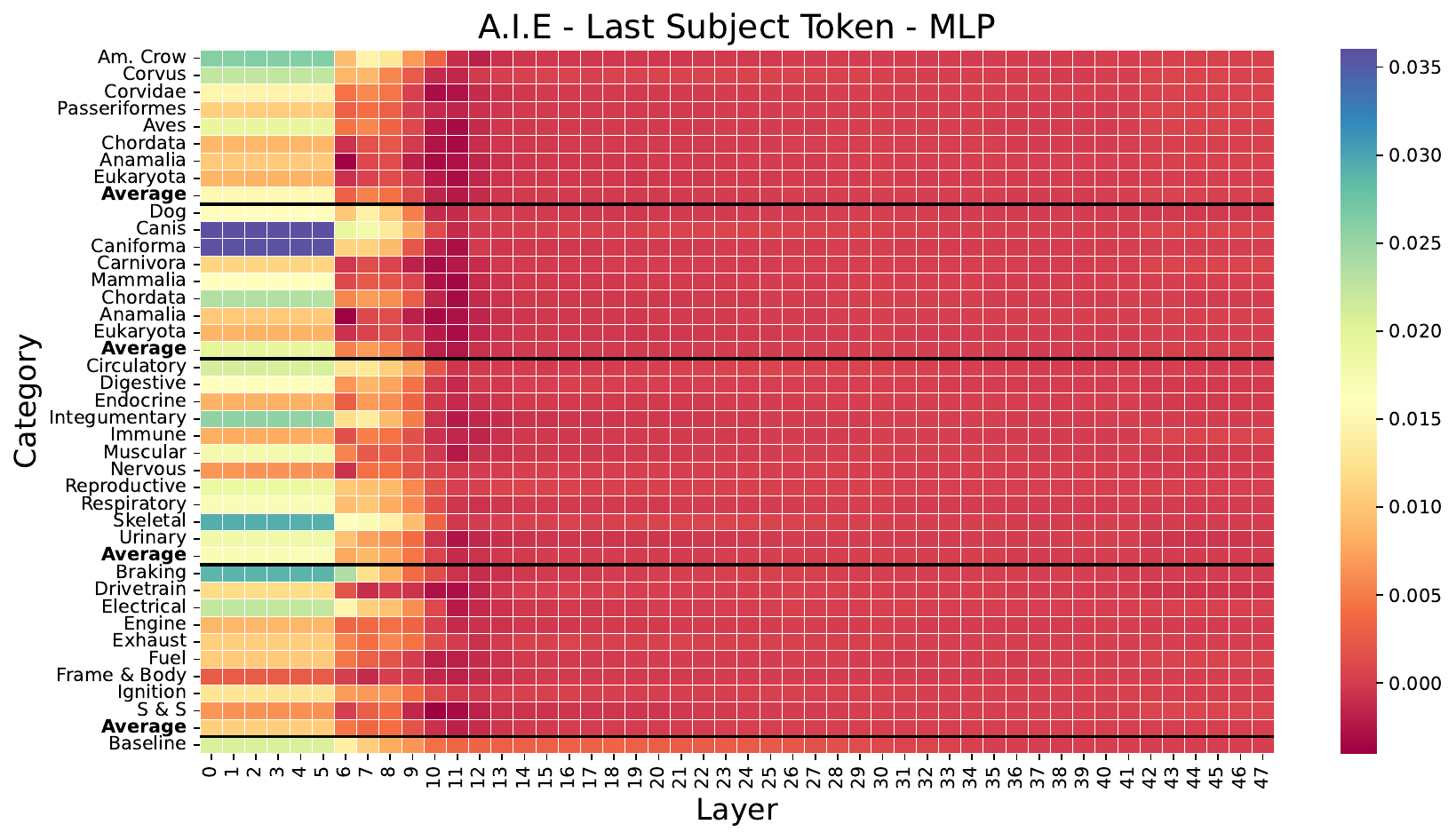}}

\caption{GPT-2-XL A.I.E Heat Map - Last Subject Token - MLP}
\label{fig:AIE-XL-MLP}
\end{figure*}

\begin{figure}[b]
    \centering
    \begin{minipage}{0.48\textwidth}
        \centering
        \includegraphics[width=1.0\textwidth]{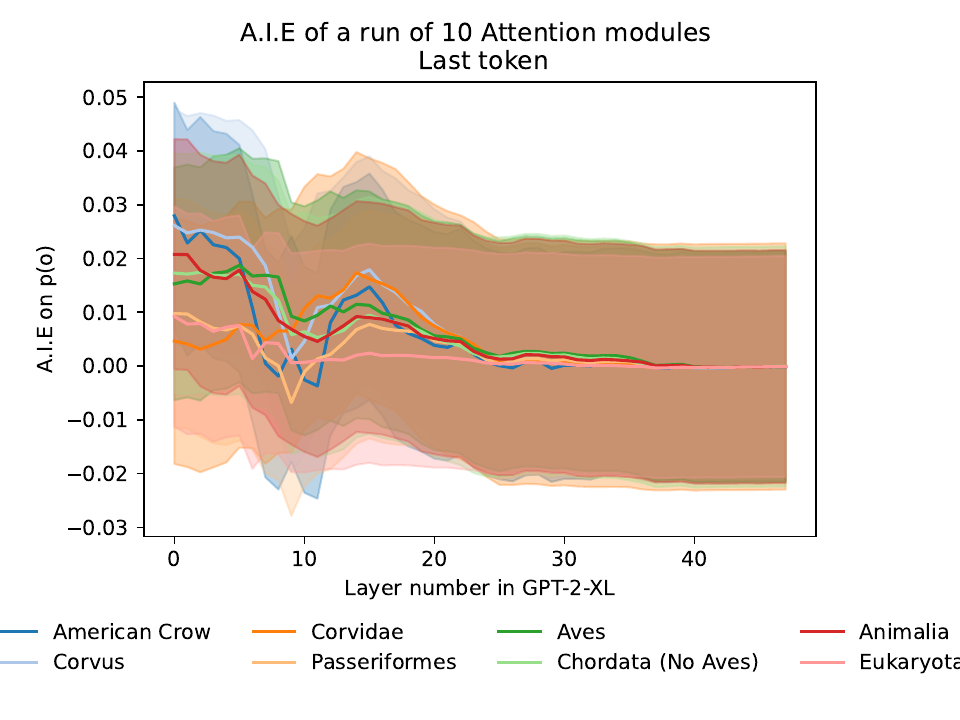} 
    \end{minipage}\hfill
    \begin{minipage}{0.48\textwidth}
        \centering
        \includegraphics[width=1.0\textwidth]{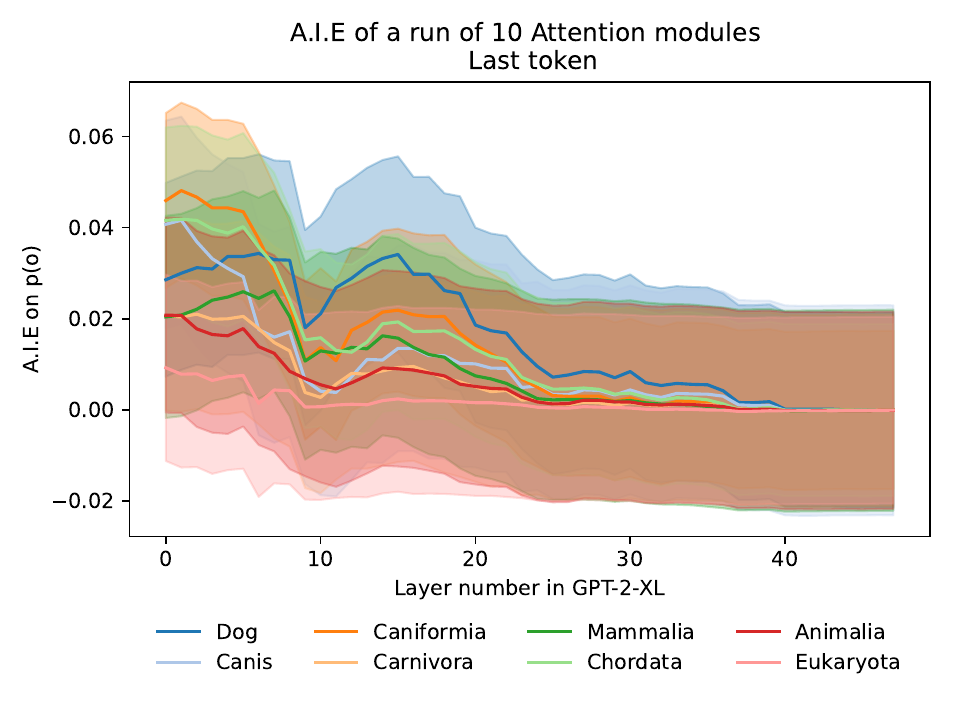} 
    \end{minipage}
    \begin{minipage}{0.48\textwidth}
        \centering
        \includegraphics[width=1.0\textwidth]{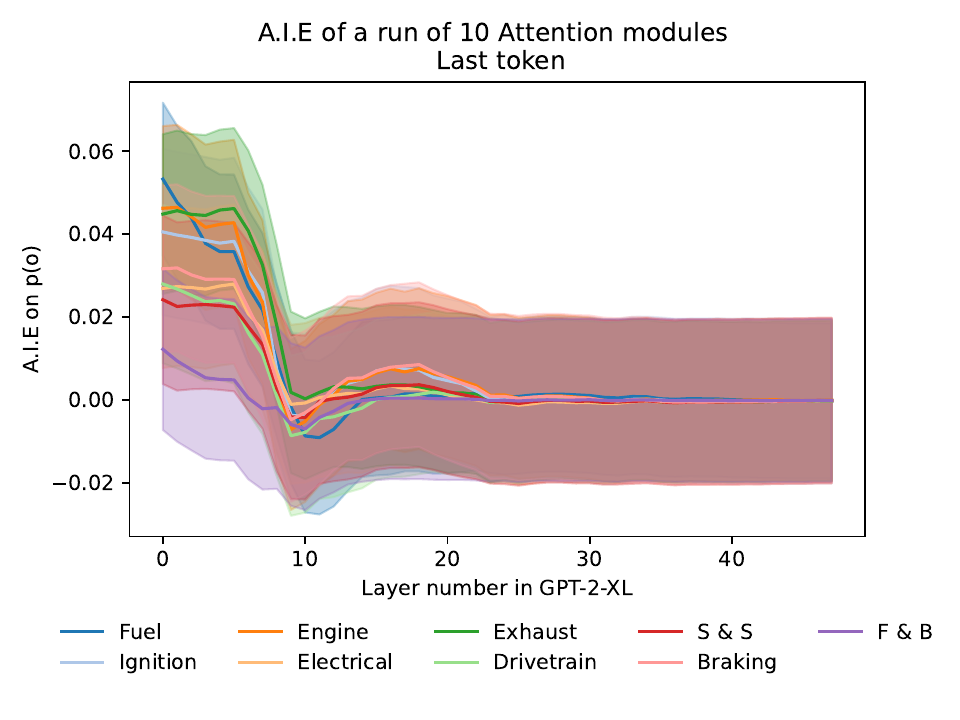} 
    \end{minipage}\hfill
    \begin{minipage}{0.48\textwidth}
        \centering
        \includegraphics[width=1.0\textwidth]{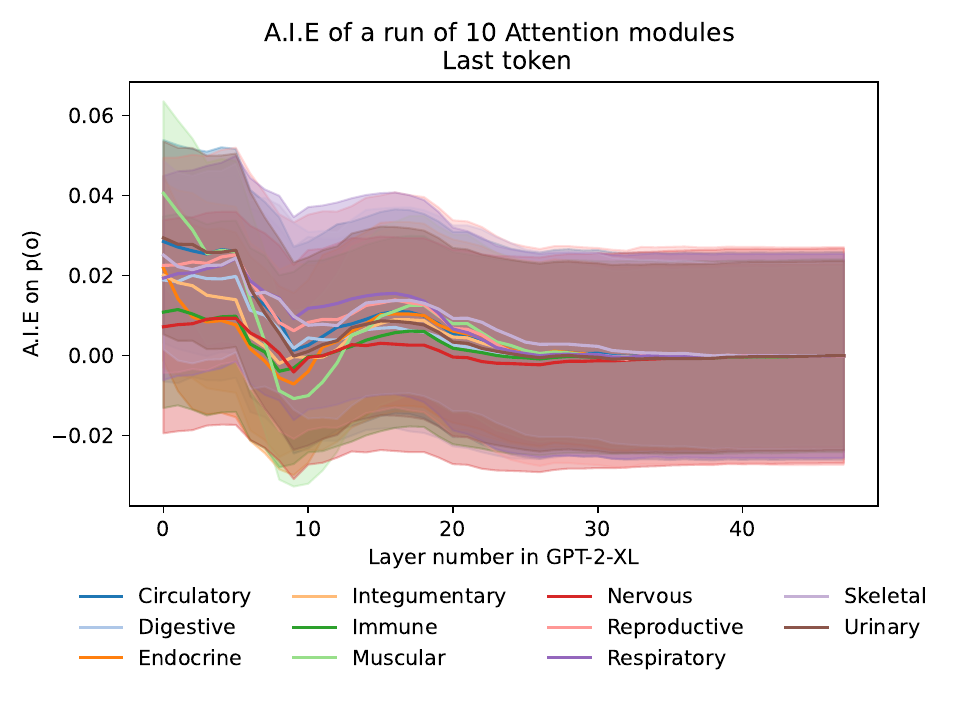}
    \end{minipage}

\caption{GPT-2-XL Line Plots - Last Token - Attention }
\label{fig:GPT2XL_Lineplots_Attn}
\end{figure}

\begin{figure*}[b]
  \centering
  \centerline{\includegraphics[width=\linewidth]{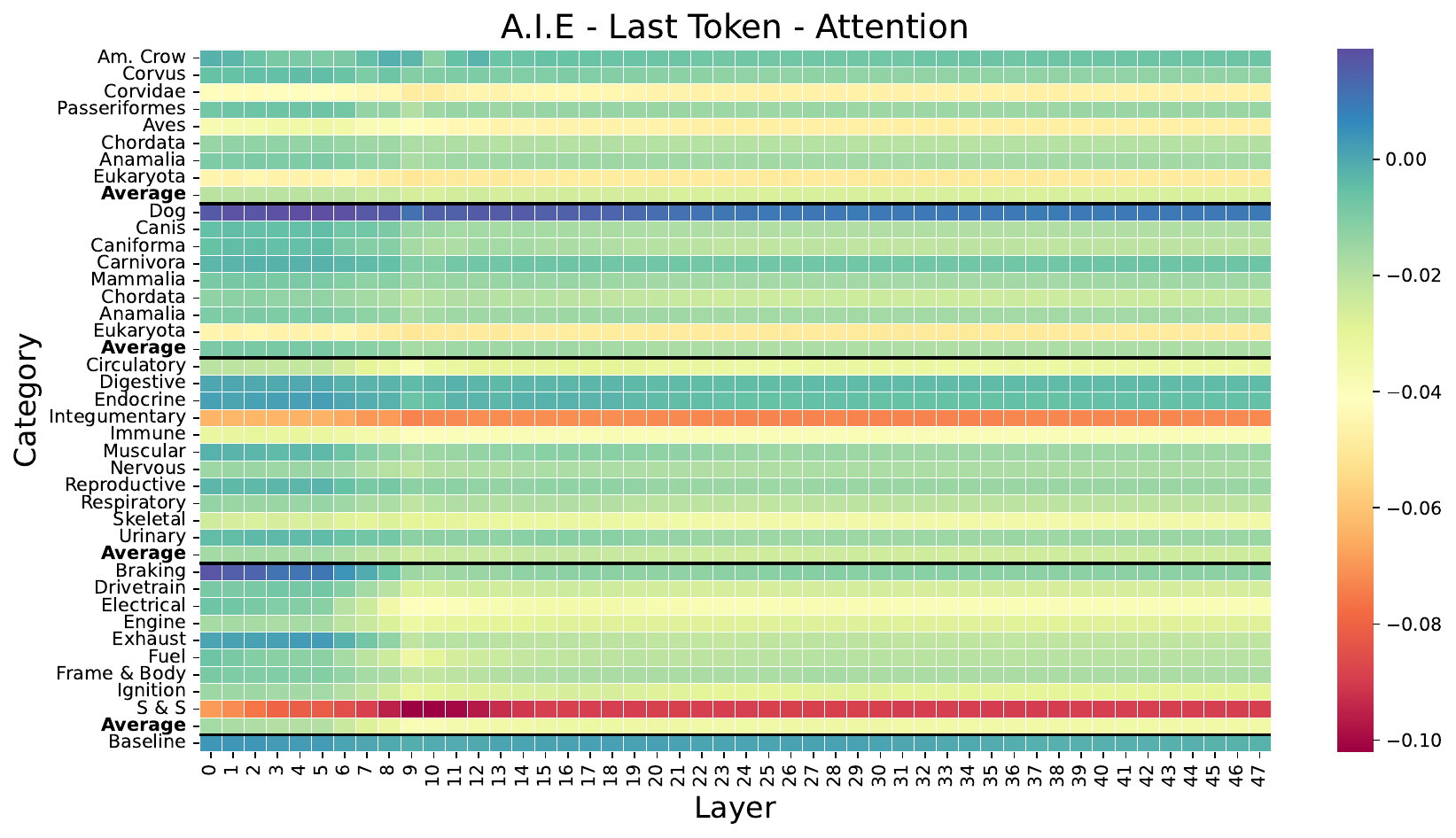}}

\caption{GPT-2-XL A.I.E Heat Map - Last Token - Attention}
\label{fig:AIE-XL-Attn}
\end{figure*}

\clearpage
\subsection{GPT-2-Large}

\renewcommand{\tabcolsep}{0.5pt}
\begin{table*}[h]
\centering
\footnotesize
\begin{tabular}{lcc|cc|lcc|cc}
\toprule

\multicolumn{1}{c}{\textbf{Category}} & \multicolumn{2}{c}{\textbf{Last Subject Token (MLP)}} & \multicolumn{2}{c}{\textbf{Last Token (ATN)}}&\multicolumn{1}{c}{\textbf{Category}} & \multicolumn{2}{c}{\textbf{Last Subject Token (MLP)}} & \multicolumn{2}{c}{\textbf{Last Token (ATN)}}\\

\cmidrule(lr){1-1}\cmidrule(lr){2-3}\cmidrule(lr){4-5}\cmidrule(lr){6-6}\cmidrule(lr){7-8}\cmidrule(lr){9-10}
\multicolumn{1}{c}{\textbf{}} &\multicolumn{1}{c}{\textbf{Max at Layer}} & \textbf{A.I.E} & \multicolumn{1}{c}{\textbf{Max at Layer}} & \textbf{A.I.E}&\multicolumn{1}{c}{\textbf{}} &\multicolumn{1}{c}{\textbf{Max at Layer}} & \textbf{A.I.E} & \multicolumn{1}{c}{\textbf{Max at Layer}} & \textbf{A.I.E} \\
\cmidrule(lr){2-2}\cmidrule(lr){5-5}\cmidrule(lr){3-3}\cmidrule(lr){4-4}\cmidrule(lr){7-7}\cmidrule(lr){8-8}\cmidrule(lr){9-9}\cmidrule(lr){10-10}

Am. Crow & 5 & 0.093 & 21  & 0.095 &  Dog & 10 & 0.128 & 26  & 0.132 \\
Corvus & 5 &  0.067 & 21 & 0.080 &  Canis & 5 & 0.070 & 22  & 0.089   \\
Corvidae & 9 & 0.037 & 22 & 0.60 & Caniformia & 5 & 0.146 & 21  & 0.135  \\
Passeriformes & 5 & 0.053 & 21 & 0.062 & Carnivora & 5 & 0.055 & 22  & 0.068  \\
Aves & 5 & 0.057  & 21 & 0.089 & Mammalia & 9 & 0.101 & 21  & 0.100  \\
Chordata* & 5 & 0.067 & 21 & 0.076 & Chordata* & 5 & 0.126 & 21  & 0.119  \\
Anamalia  & 5 & 0.079 & 21 & 0.083 &  Anamalia & 5 & 0.079 & 21 & 0.083 \\
Eukaryota & 5 & 0.057 & 22 & 0.060 & Eukaryota  & 5 & 0.057 & 22 & 0.060  \\
\textbf{Baseline} &5  & 0.061 & 26 & 0.130 & \textbf{Baseline} &5  & 0.061 & 26 & 0.130  \\

\bottomrule
\end{tabular}
\caption{Taxonomic Statistics - GPT-2-Large}
\label{tab:taxonomicGPT2L}
\vspace{-10pt}
\end{table*}

\renewcommand{\tabcolsep}{0.5pt}
\begin{table*}[h]
\centering
\footnotesize
\begin{tabular}{lcc|cc|lcc|cc}
\toprule

\multicolumn{1}{c}{\textbf{Category}} & \multicolumn{2}{c}{\textbf{Last Subject Token (MLP)}} & \multicolumn{2}{c}{\textbf{Last Token (ATN)}}&\multicolumn{1}{c}{\textbf{Category}} & \multicolumn{2}{c}{\textbf{Last Subject Token (MLP)}} & \multicolumn{2}{c}{\textbf{Last Token (ATN)}}\\

\cmidrule(lr){1-1}\cmidrule(lr){2-3}\cmidrule(lr){4-5}\cmidrule(lr){6-6}\cmidrule(lr){7-8}\cmidrule(lr){9-10}
\multicolumn{1}{c}{\textbf{}} &\multicolumn{1}{c}{\textbf{Max at Layer}} & \textbf{A.I.E} & \multicolumn{1}{c}{\textbf{Max at Layer}} & \textbf{A.I.E}&\multicolumn{1}{c}{\textbf{}} &\multicolumn{1}{c}{\textbf{Max at Layer}} & \textbf{A.I.E} & \multicolumn{1}{c}{\textbf{Max at Layer}} & \textbf{A.I.E} \\
\cmidrule(lr){2-2}\cmidrule(lr){5-5}\cmidrule(lr){3-3}\cmidrule(lr){4-4}\cmidrule(lr){7-7}\cmidrule(lr){8-8}\cmidrule(lr){9-9}\cmidrule(lr){10-10}

Circulatory & 5 & 0.076 & 22  & 0.121 & Braking & 5 & 0.058 & 21  & 0.088 \\
Digestive & 5 &  0.079 & 22 & 0.117 &  Drivetrain & 5 & 0.062 & 21  & 0.081   \\
Endocrine & 5 & 0.088 & 22 & 0.110 & Electrical & 5 & 0.044 & 22  & 0.081  \\
Integumentary & 5 & 0.066 & 21 & 0.078 & Engine & 5 & 0.078 & 21  & 0.115  \\
Immune & 5 & 0.049  & 22 & 0.082 & Exhaust & 5 & 0.048 & 22  & 0.077  \\
Muscular & 5 & 0.094 & 21 & 0.123 & Fuel & 5 & 0.067 & 22  & 0.068  \\
Nervous & 5 & 0.037 & 22 & 0.068 & Frame  & \multirow{2}{*}{4} & \multirow{2}{*}{0.038} & \multirow{2}{*}{22}  & \multirow{2}{*}{0.036}   \\
Reproductive & 5 & 0.060 & 22 & 0.111 & \& Body & &  &   &  \\
Respiratory & 5 & 0.057 & 22 & 0.111 & Ignition  & 5 & 0.063 & 22  & 0.090  \\
Skeletal & 5 & 0.083 & 21 & 0.115 & Suspension & \multirow{2}{*}{5} & \multirow{2}{*}{0.055} & \multirow{2}{*}{21}  & \multirow{2}{*}{0.085}  \\
Urinary & 5 & 0.089 & 22 & 0.129 & \& Steering &  & &  &  \\
\textbf{Baseline} &5  & 0.061 & 26 & 0.130 & \textbf{Baseline} &5  & 0.061 & 26 & 0.130   \\

\bottomrule
\end{tabular}
\caption{Meronomic Statistics - GPT-2-Large}
\label{tab:meronomic_statsGPT2L}
\vspace{-10pt}
\end{table*}

\begin{figure*}[]
  \centering
  \centerline{\includegraphics[width=\linewidth]{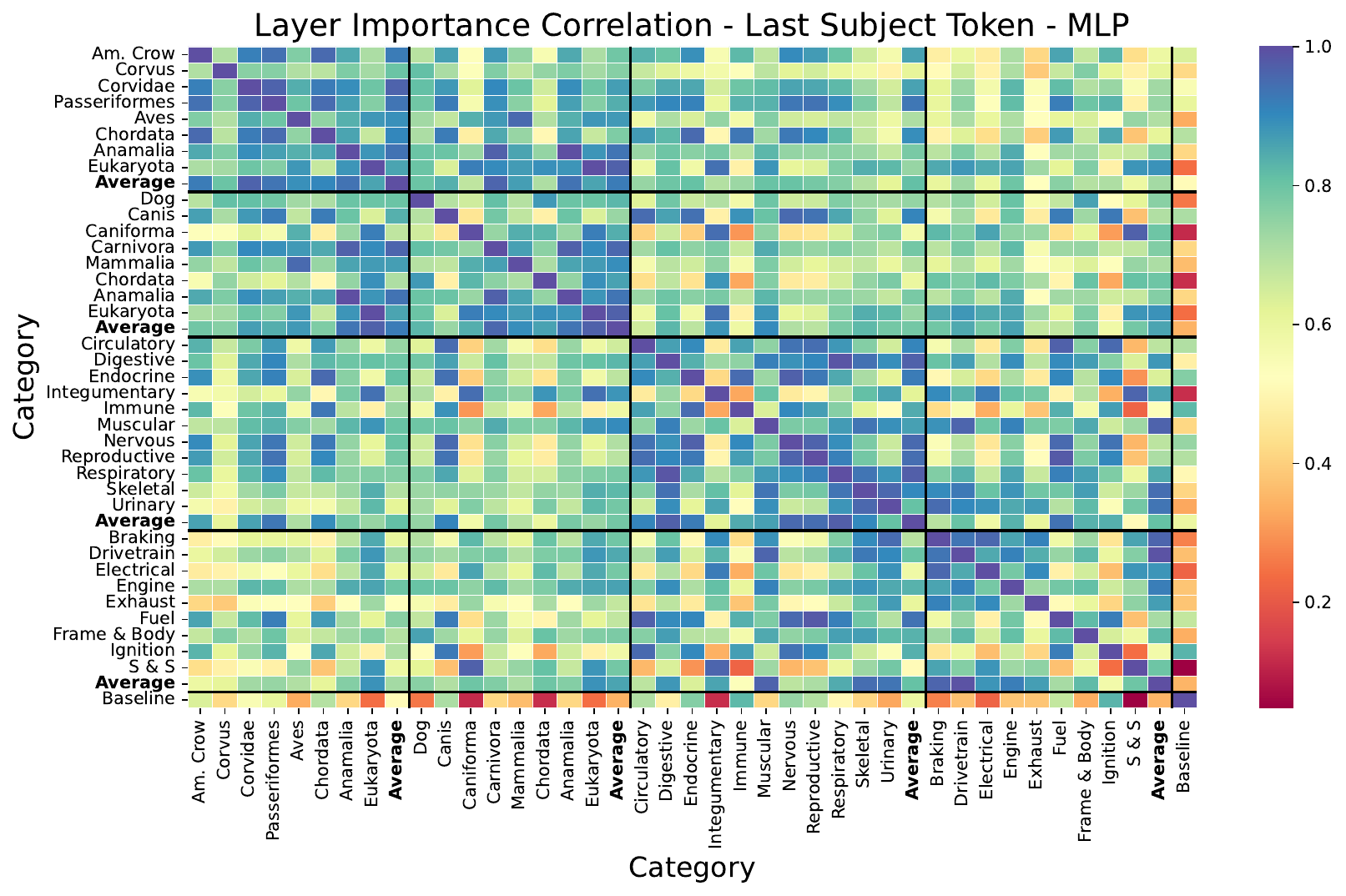}}

\caption{GPT-2-L Layer Importance Heatmap - MLP}
\label{fig:LIH-L-MLP}

\end{figure*}

\begin{figure*}[]
  \centering
  \centerline{\includegraphics[width=\linewidth]{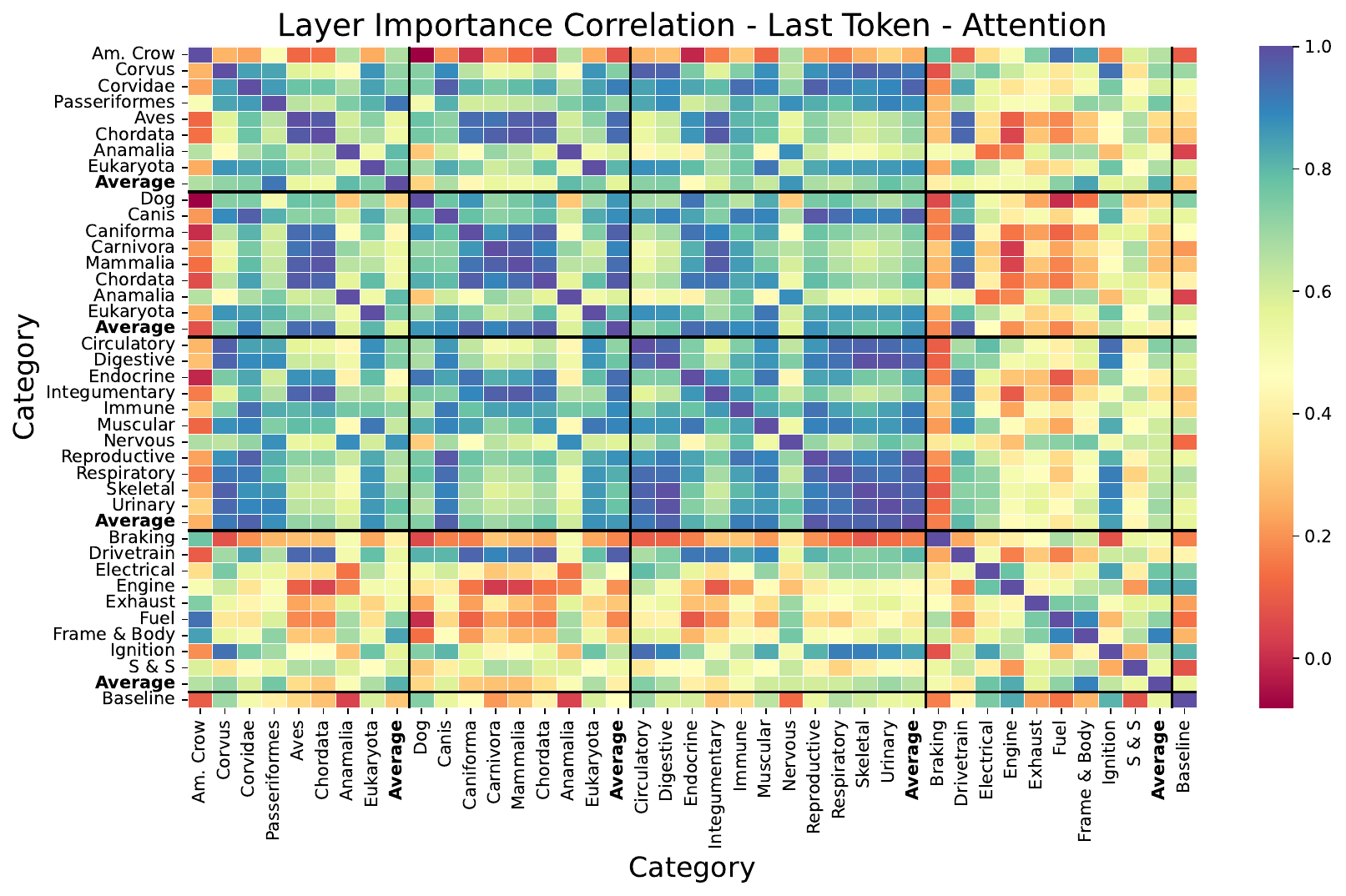}}

\caption{GPT-2-L Layer Importance Heatmap - Attention}
\label{fig:LIH-L-Attn}

\end{figure*}

\begin{figure}
    \centering
    \begin{minipage}{0.48\textwidth}
        \centering
        \includegraphics[width=1.0\textwidth]{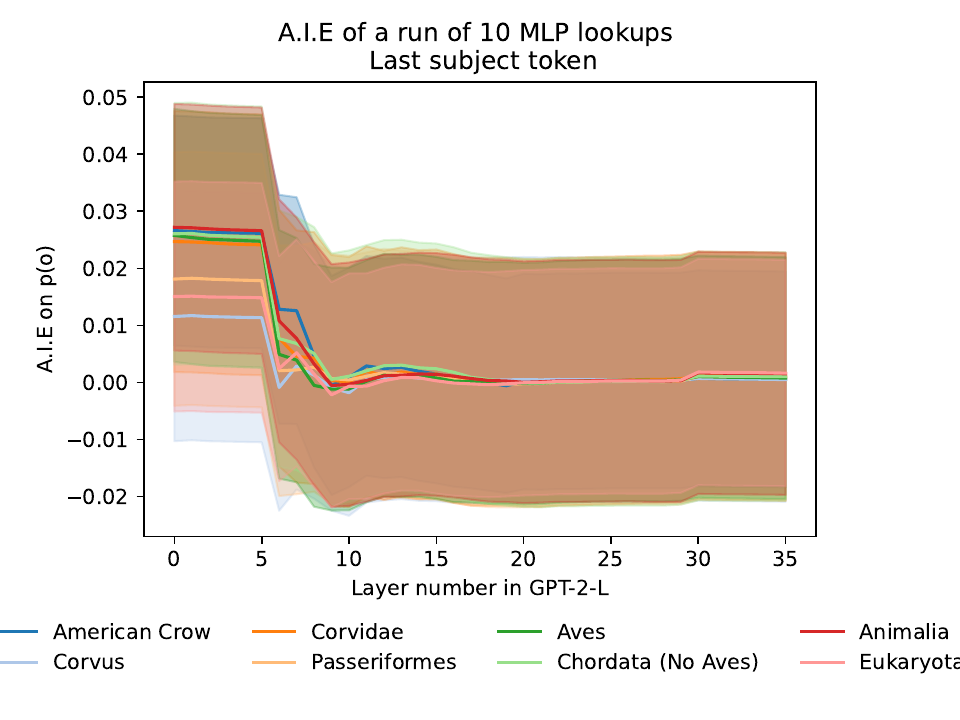} 
    \end{minipage}\hfill
    \begin{minipage}{0.48\textwidth}
        \centering
        \includegraphics[width=1.0\textwidth]{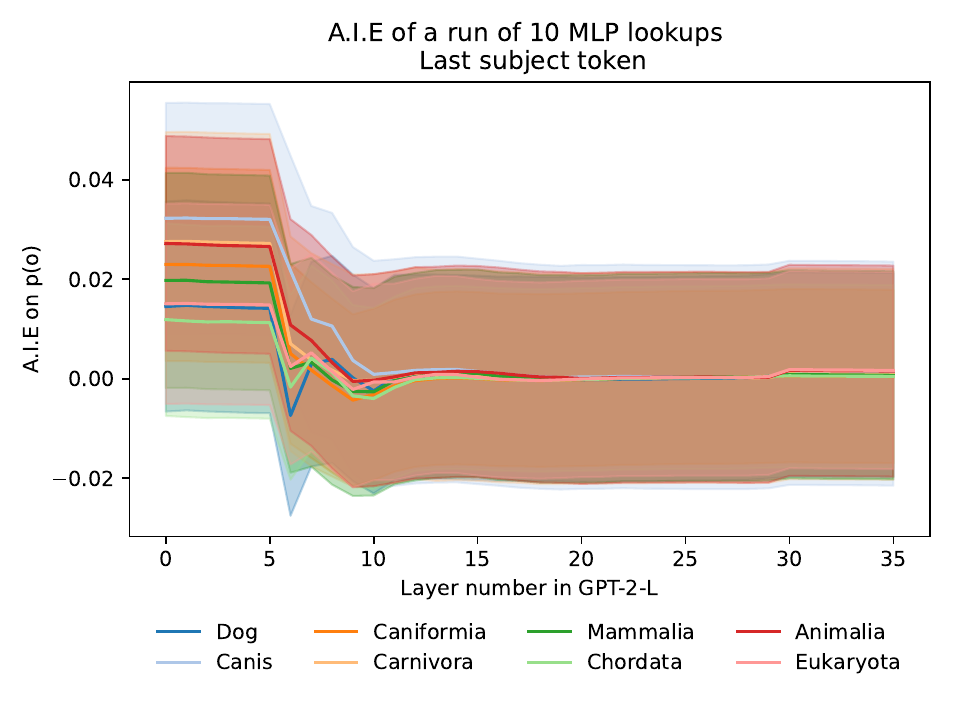} 
    \end{minipage}
    \begin{minipage}{0.48\textwidth}
        \centering
        \includegraphics[width=1.0\textwidth]{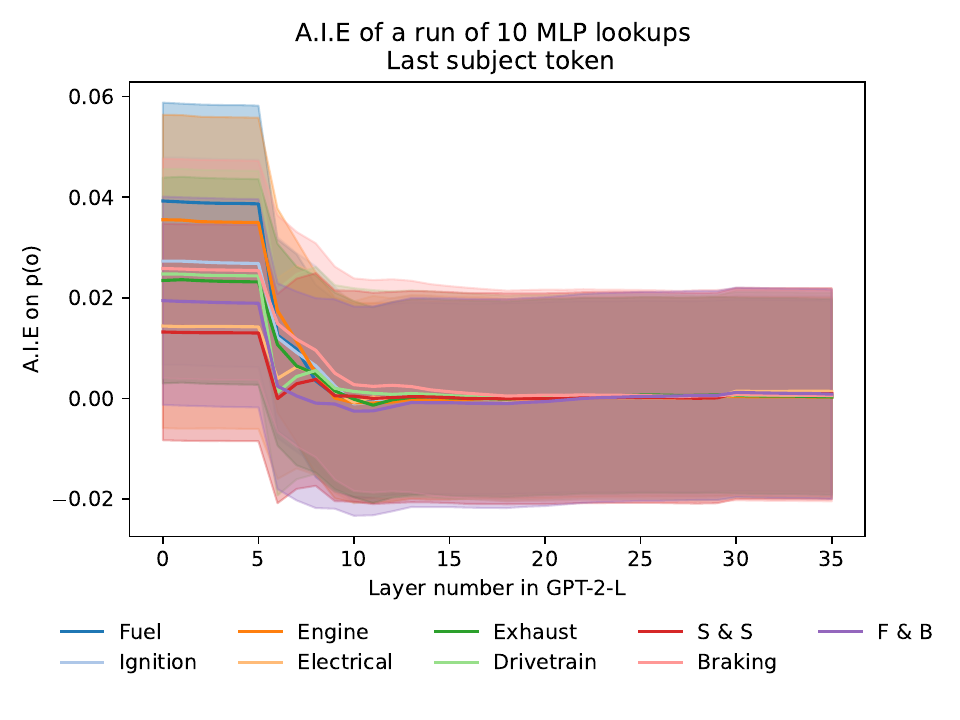} 
    \end{minipage}\hfill
    \begin{minipage}{0.48\textwidth}
        \centering
        \includegraphics[width=1.0\textwidth]{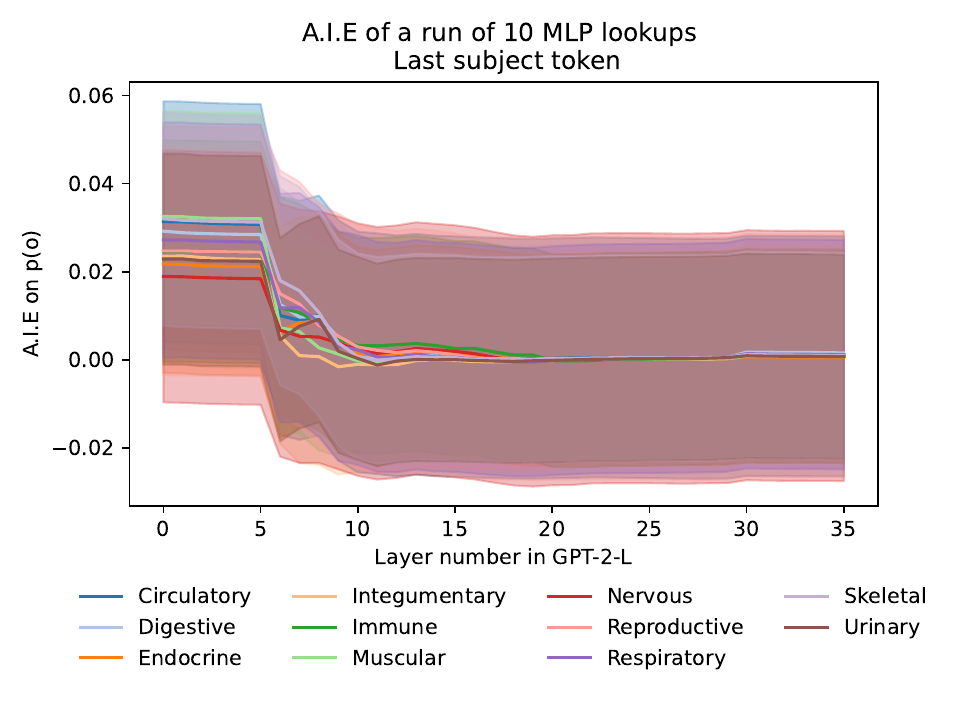} 
    \end{minipage}

\caption{GPT-2-L Line Plots - Last Subject Token - MLP }
\label{fig:GPT2L_Lineplots_MLP}
\end{figure}

\begin{figure*}[]
  \centering
  \centerline{\includegraphics[width=\linewidth]{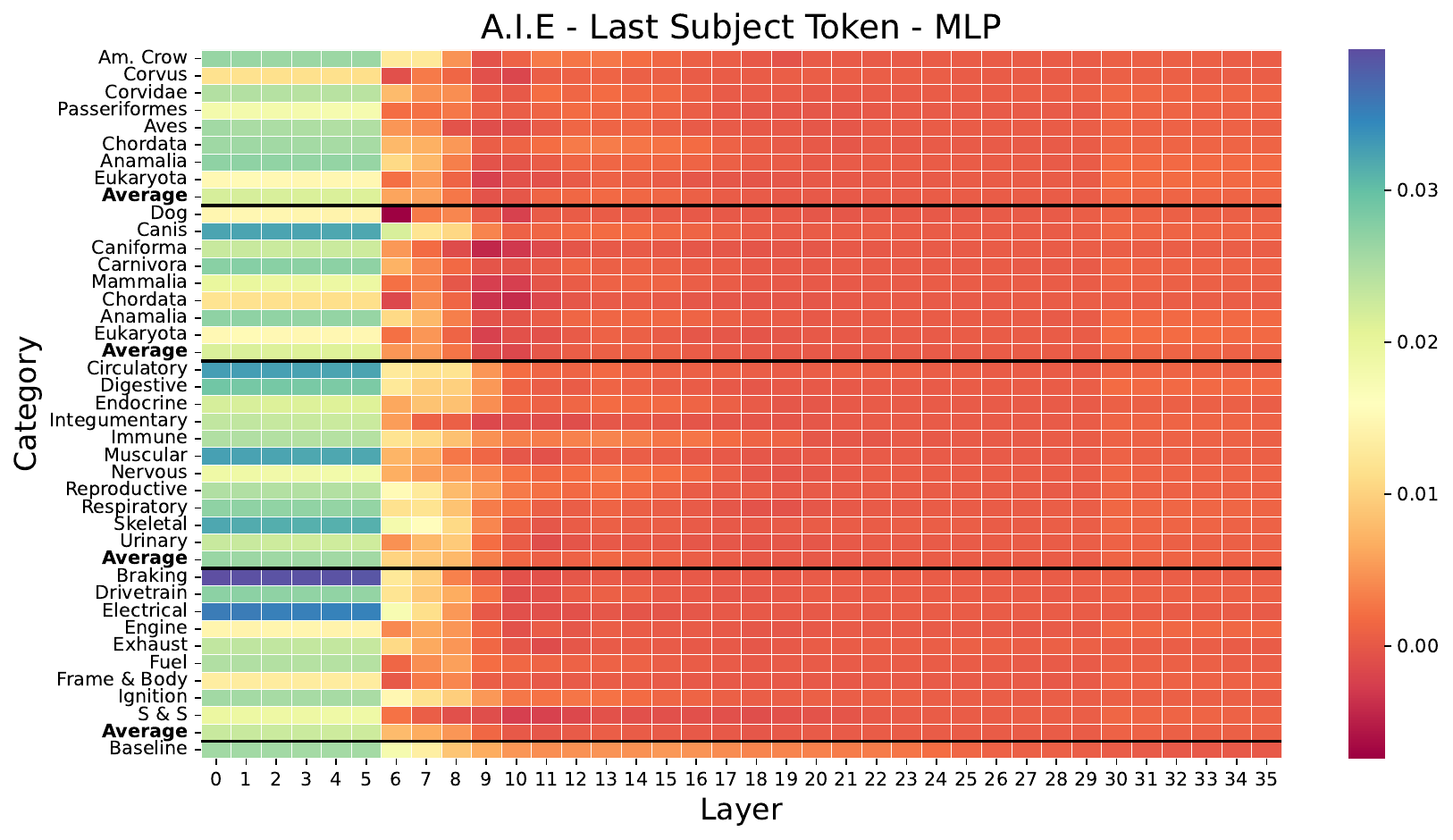}}

\caption{GPT-2-L A.I.E Heat Map - Last Subject Token - MLP}
\label{fig:LIH-L-MLP}

\end{figure*}

\begin{figure}
    \centering
    \begin{minipage}{0.48\textwidth}
        \centering
        \includegraphics[width=1.0\textwidth]{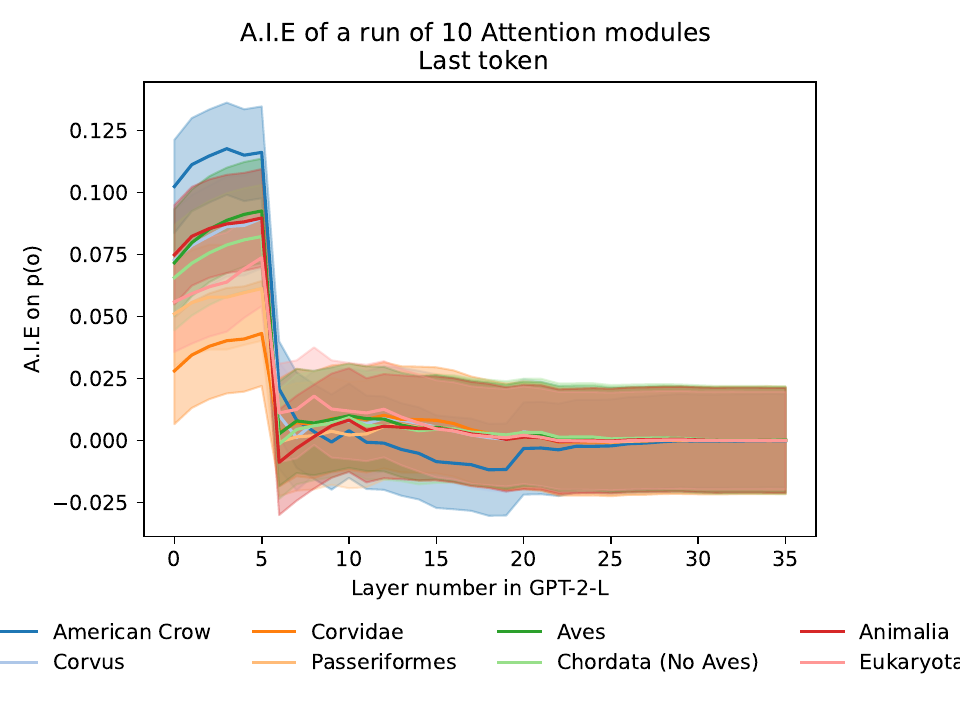}
    \end{minipage}\hfill
    \begin{minipage}{0.48\textwidth}
        \centering
        \includegraphics[width=1.0\textwidth]{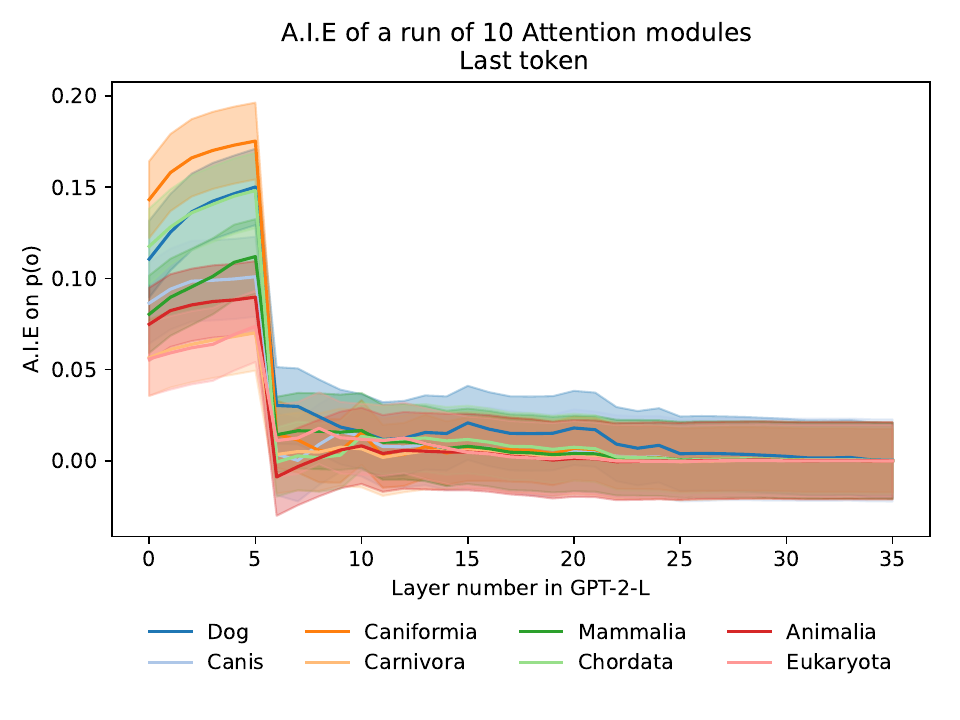} 
    \end{minipage}
    \begin{minipage}{0.48\textwidth}
        \centering
        \includegraphics[width=1.0\textwidth]{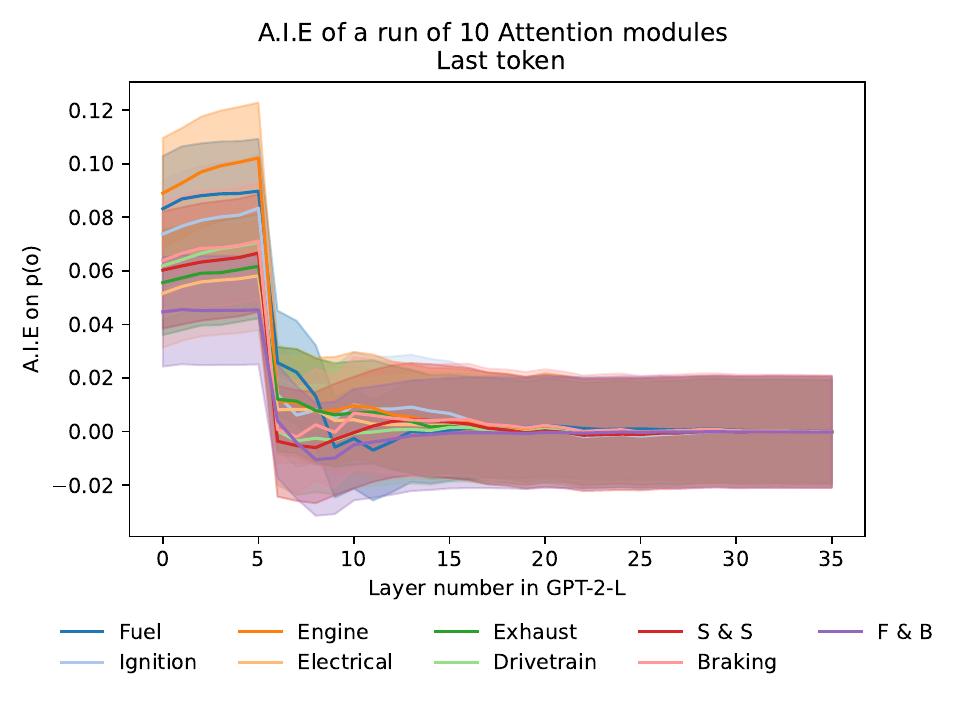} 
    \end{minipage}\hfill
    \begin{minipage}{0.48\textwidth}
        \centering
        \includegraphics[width=1.0\textwidth]{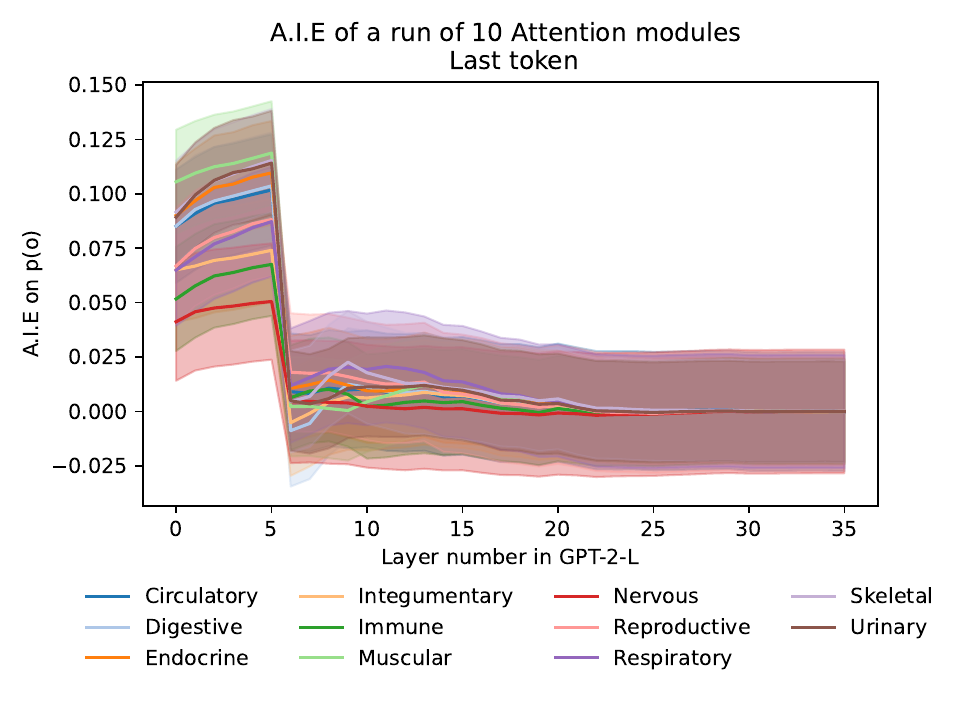} 
    \end{minipage}

\caption{GPT-2-L Line Plots - Last Token - Attention }
\label{fig:GPT2L_Lineplots_Attn}
\end{figure}

\begin{figure*}[]
  \centering
  \centerline{\includegraphics[width=\linewidth]{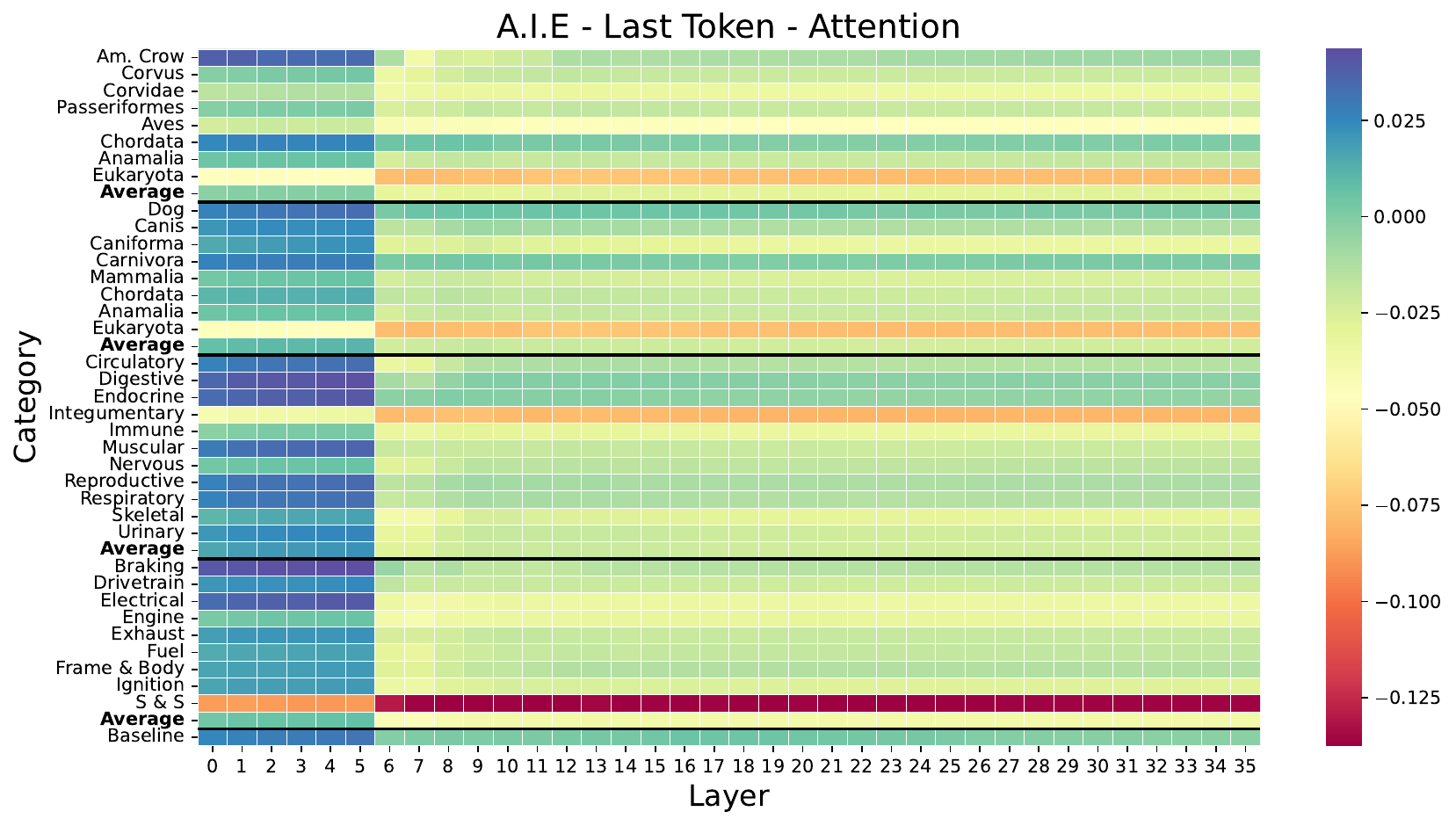}}

\caption{GPT-2-L A.I.E Heat Map - Last Token - Attention}
\label{fig:LIH-L-Attn}

\end{figure*}

\clearpage

\subsection{GPT-2-Medium}
\renewcommand{\tabcolsep}{0.5pt}
\begin{table*}[h]
\centering
\footnotesize
\begin{tabular}{lcc|cc|lcc|cc}
\toprule

\multicolumn{1}{c}{\textbf{Category}} & \multicolumn{2}{c}{\textbf{Last Subject Token (MLP)}} & \multicolumn{2}{c}{\textbf{Last Token (ATN)}}&\multicolumn{1}{c}{\textbf{Category}} & \multicolumn{2}{c}{\textbf{Last Subject Token (MLP)}} & \multicolumn{2}{c}{\textbf{Last Token (ATN)}}\\

\cmidrule(lr){1-1}\cmidrule(lr){2-3}\cmidrule(lr){4-5}\cmidrule(lr){6-6}\cmidrule(lr){7-8}\cmidrule(lr){9-10}
\multicolumn{1}{c}{\textbf{}} &\multicolumn{1}{c}{\textbf{Max at Layer}} & \textbf{A.I.E} & \multicolumn{1}{c}{\textbf{Max at Layer}} & \textbf{A.I.E}&\multicolumn{1}{c}{\textbf{}} &\multicolumn{1}{c}{\textbf{Max at Layer}} & \textbf{A.I.E} & \multicolumn{1}{c}{\textbf{Max at Layer}} & \textbf{A.I.E} \\
\cmidrule(lr){2-2}\cmidrule(lr){5-5}\cmidrule(lr){3-3}\cmidrule(lr){4-4}\cmidrule(lr){7-7}\cmidrule(lr){8-8}\cmidrule(lr){9-9}\cmidrule(lr){10-10}

Am. Crow & 3 & 0.126 & 15  & 0.118 &  Dog & 5 & 0.121 & 19  & 0.200 \\
Corvus & 5 &  0.085 & 19 & 0.120 &  Canis & 4 & 0.099 & 10  & 0.137   \\
Corvidae & 4 & 0.046 & 19 & 0.081 & Caniformia & 3 & 0.135 & 13  & 0.165  \\
Passeriformes & 3 & 0.054 & 19 & 0.092 & Carnivora & 3 & 0.057 & 19  & 0.108  \\
Aves & 3 & 0.084  & 19 & 0.120 & Mammalia & 3 & 0.093 & 19  & 0.132  \\
Chordata* & 3 & 0.063 & 19 & 0.115 & Chordata* & 4 & 0.125 & 19  & 0.154  \\
Anamalia  & 3 & 0.093 & 19 & 0.126 &  Anamalia & 3 & 0.093 & 19 & 0.126  \\
Eukaryota & 3 & 0.052 & 19 & 0.078 & Eukaryota  & 3 & 0.052 & 19 & 0.078  \\
\textbf{Baseline} & 5 & 0.075 & 19 & 0.163  &  \textbf{Baseline} & 5 & 0.075 & 19 & 0.163  \\

\bottomrule
\end{tabular}
\caption{Taxonomic Statistics - GPT-2-Medium}
\label{tab:taxonomicGPT2M}
\vspace{-10pt}
\end{table*}

\renewcommand{\tabcolsep}{0.5pt}
\begin{table*}[h]
\centering
\footnotesize
\begin{tabular}{lcc|cc|lcc|cc}
\toprule

\multicolumn{1}{c}{\textbf{Category}} & \multicolumn{2}{c}{\textbf{Last Subject Token (MLP)}} & \multicolumn{2}{c}{\textbf{Last Token (ATN)}}&\multicolumn{1}{c}{\textbf{Category}} & \multicolumn{2}{c}{\textbf{Last Subject Token (MLP)}} & \multicolumn{2}{c}{\textbf{Last Token (ATN)}}\\

\cmidrule(lr){1-1}\cmidrule(lr){2-3}\cmidrule(lr){4-5}\cmidrule(lr){6-6}\cmidrule(lr){7-8}\cmidrule(lr){9-10}
\multicolumn{1}{c}{\textbf{}} &\multicolumn{1}{c}{\textbf{Max at Layer}} & \textbf{A.I.E} & \multicolumn{1}{c}{\textbf{Max at Layer}} & \textbf{A.I.E}&\multicolumn{1}{c}{\textbf{}} &\multicolumn{1}{c}{\textbf{Max at Layer}} & \textbf{A.I.E} & \multicolumn{1}{c}{\textbf{Max at Layer}} & \textbf{A.I.E} \\
\cmidrule(lr){2-2}\cmidrule(lr){5-5}\cmidrule(lr){3-3}\cmidrule(lr){4-4}\cmidrule(lr){7-7}\cmidrule(lr){8-8}\cmidrule(lr){9-9}\cmidrule(lr){10-10}

Circulatory & 3 & 0.068 & 19  & 0.139 & Braking & 2 & 0.053 & 19  & 0.097 \\
Digestive & 5 &  0.060 & 19 & 0.108 &  Drivetrain & 3 & 0.050 & 19  & 0.106  \\
Endocrine & 5 & 0.087 & 19 & 0.135 & Electrical & 5 & 0.043 & 17  & 0.110  \\
Integumentary & 5 & 0.079 & 19 & 0.122 & Engine & 5 & 0.073 & 17  & 0.136  \\
Immune & 5 & 0.047  & 19 & 0.090 & Exhaust & 5 & 0.042 & 19  & 0.100  \\
Muscular & 4 & 0.092 & 15 & 0.143 & Fuel & 3 & 0.059 & 19  & 0.097  \\
Nervous & 3 & 0.030 & 19 & 0.081 & Frame  & \multirow{2}{*}{3} & \multirow{2}{*}{0.039} & \multirow{2}{*}{19}  & \multirow{2}{*}{0.050}   \\
Reproductive & 5 & 0.052 & 19 & 0.119 & \& Body & &  &   &  \\
Respiratory & 5 & 0.057 & 19 & 0.138 & Ignition  & 3 & 0.062 & 19  & 0.104  \\
Skeletal & 5 & 0.070 & 16 & 0.119 & Suspension & \multirow{2}{*}{2} & \multirow{2}{*}{0.051} & \multirow{2}{*}{16}  & \multirow{2}{*}{0.091}  \\
Urinary & 5 & 0.093 & 16 & 0.146 & \& Steering &  & &  &  \\
\textbf{Baseline} & 5 & 0.075 & 19 & 0.163 &  \textbf{Baseline} & 5 & 0.075 & 19 & 0.163  \\

\bottomrule
\end{tabular}
\caption{Meronomic Statistics - GPT-2-Medium}
\label{tab:meronomic_statsGPT2M}
\vspace{-10pt}
\end{table*}

\begin{figure*}[]
  \centering
  \centerline{\includegraphics[width=\linewidth]{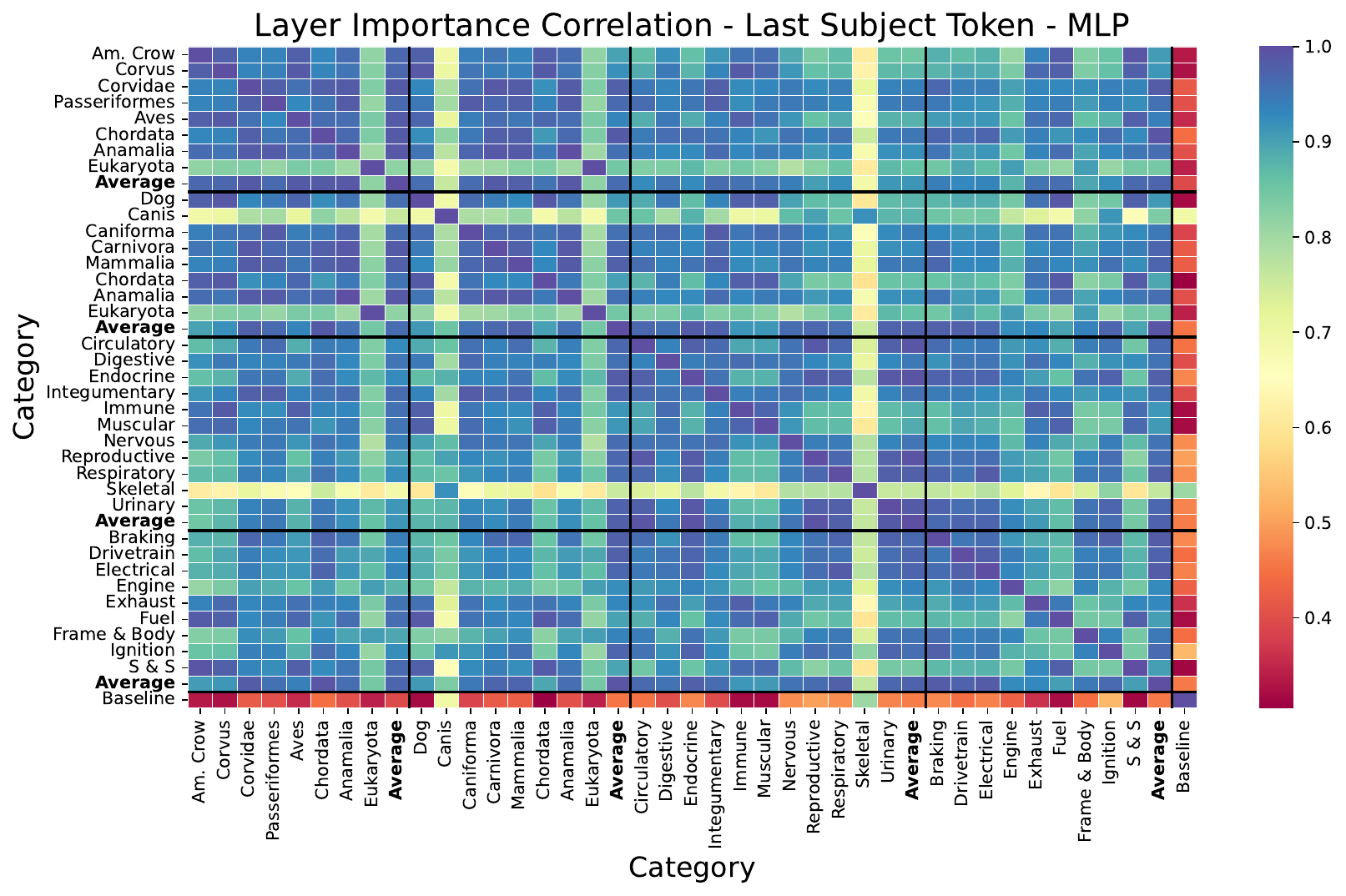}}

\caption{GPT-2-M Layer Importance Heat Map - Last Subject Token - MLP}
\label{fig:AIE-M-MLP}

\end{figure*}

\begin{figure*}[]
  \centering
  \centerline{\includegraphics[width=\linewidth]{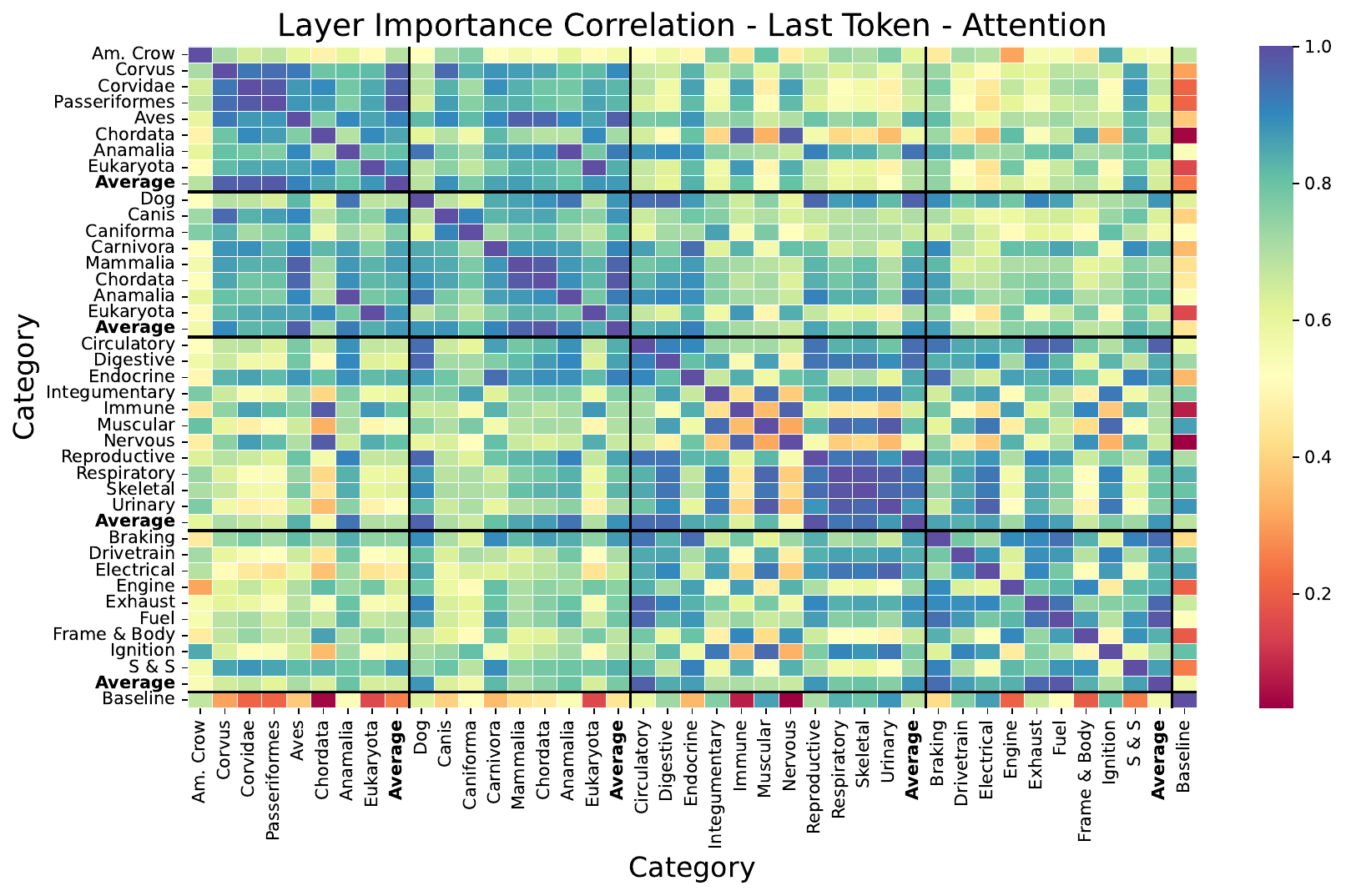}}

\caption{GPT-2-M Layer Importance Heat Map - Last Subject Token - Attention}
\label{fig:AIE-M-Attn}

\end{figure*}

\begin{figure}
    \centering
    \begin{minipage}{0.48\textwidth}
        \centering
        \includegraphics[width=1.0\textwidth]{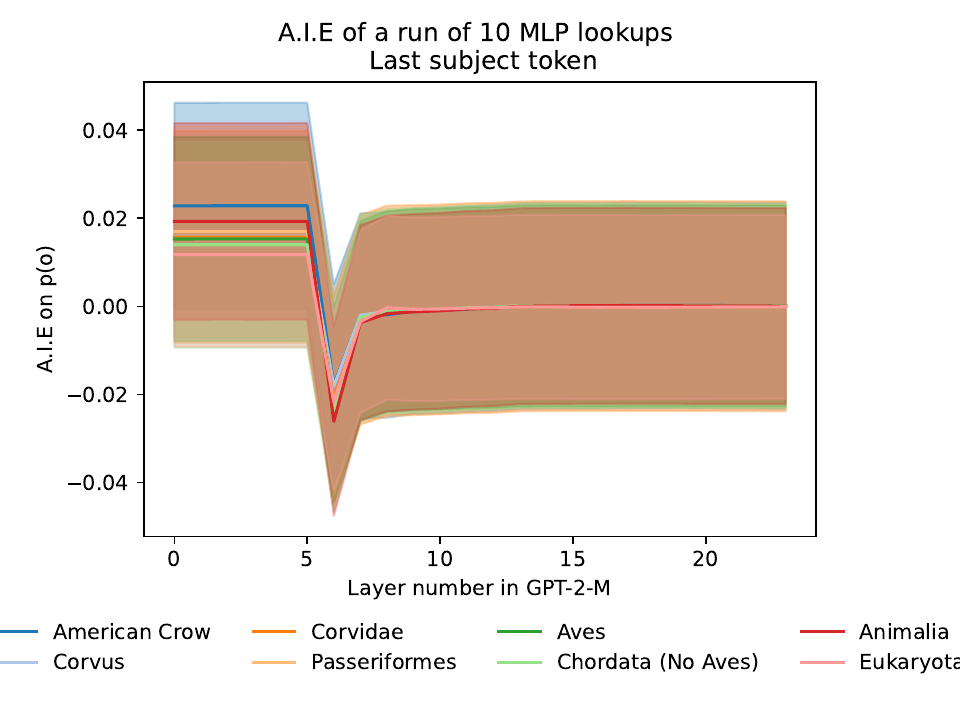} 
    \end{minipage}\hfill
    \begin{minipage}{0.48\textwidth}
        \centering
        \includegraphics[width=1.0\textwidth]{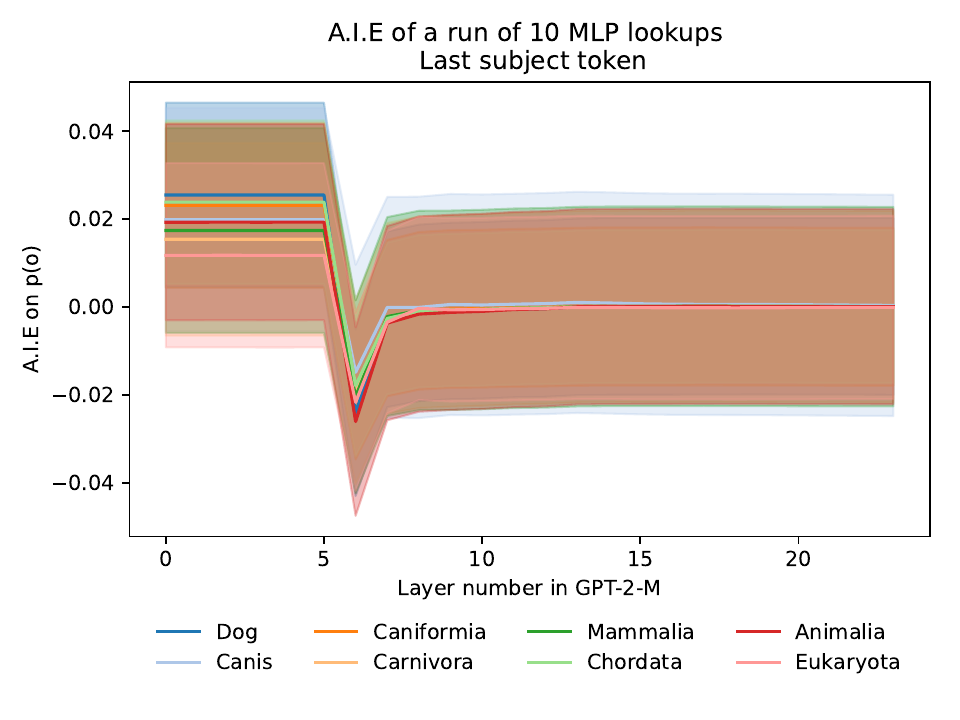}
    \end{minipage}
    \begin{minipage}{0.48\textwidth}
        \centering
        \includegraphics[width=1.0\textwidth]{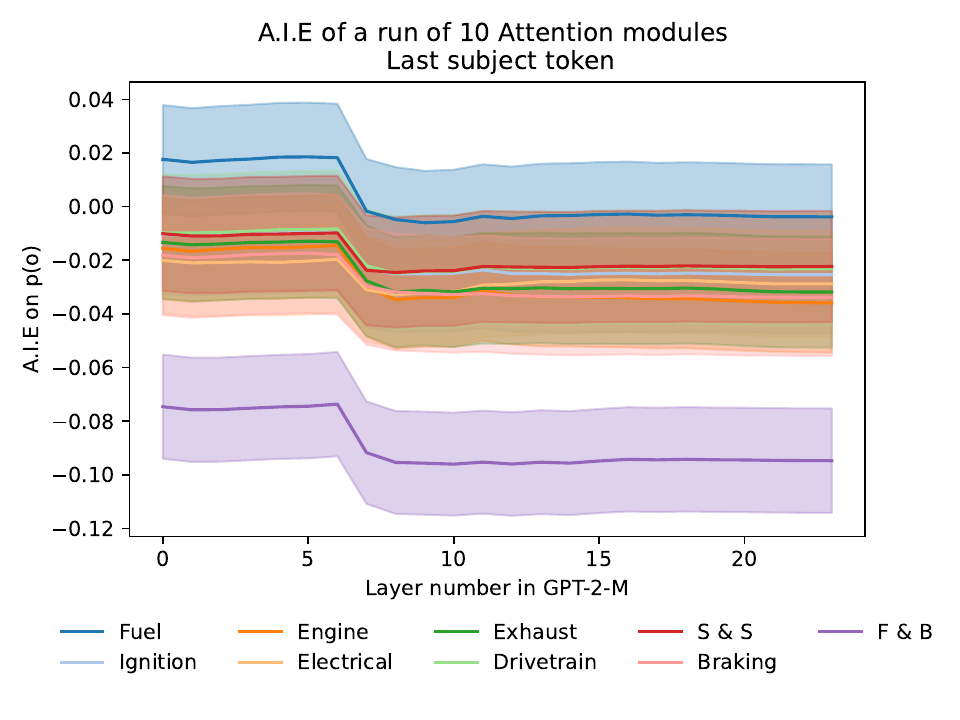} 
    \end{minipage}\hfill
    \begin{minipage}{0.48\textwidth}
        \centering
        \includegraphics[width=1.0\textwidth]{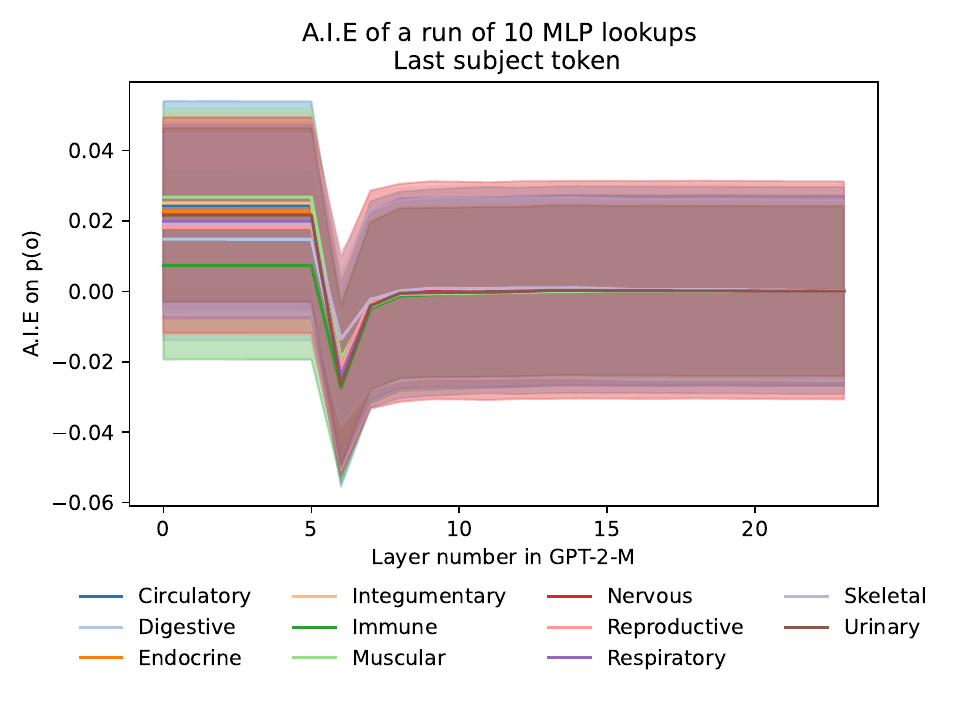} 
    \end{minipage}

\caption{GPT-2-M Line Plots - Last Subject Token - MLP }
\label{fig:GPT2M_Lineplots_MLP}
\end{figure}

\begin{figure*}[]
  \centering
  \centerline{\includegraphics[width=\linewidth]{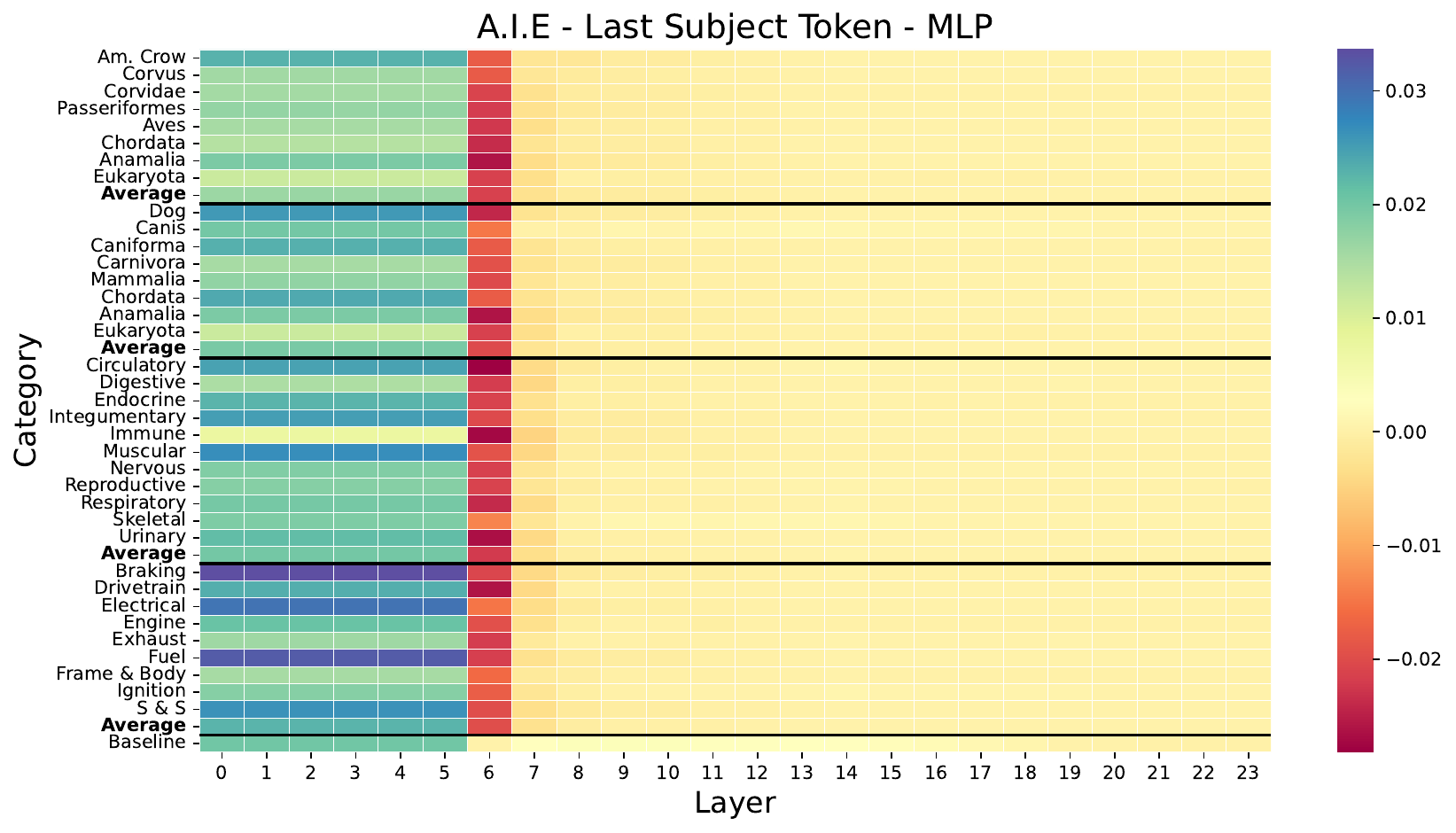}}

\caption{GPT-2-M A.I.E Heat Map - MLP}
\label{fig:LIH-M-MLP}

\end{figure*}

\begin{figure}
    \centering
    \begin{minipage}{0.48\textwidth}
        \centering
        \includegraphics[width=1.0\textwidth]{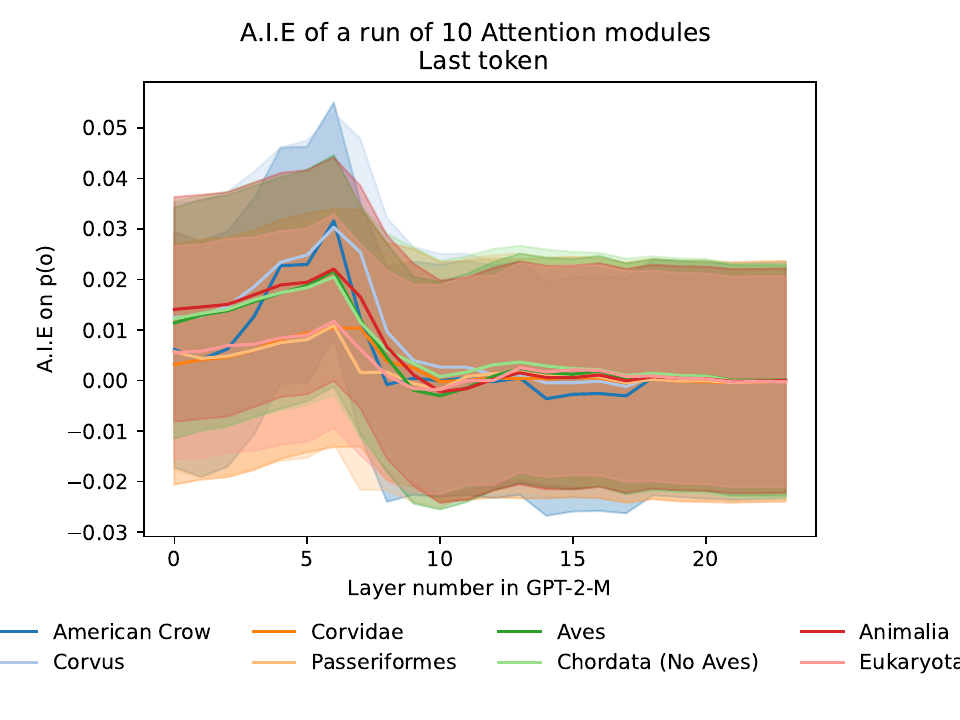} 
    \end{minipage}\hfill
    \begin{minipage}{0.48\textwidth}
        \centering
        \includegraphics[width=1.0\textwidth]{Images/m/Last_subject_token_mlp_dogs.pdf} 
    \end{minipage}
    \begin{minipage}{0.48\textwidth}
        \centering
        \includegraphics[width=1.0\textwidth]{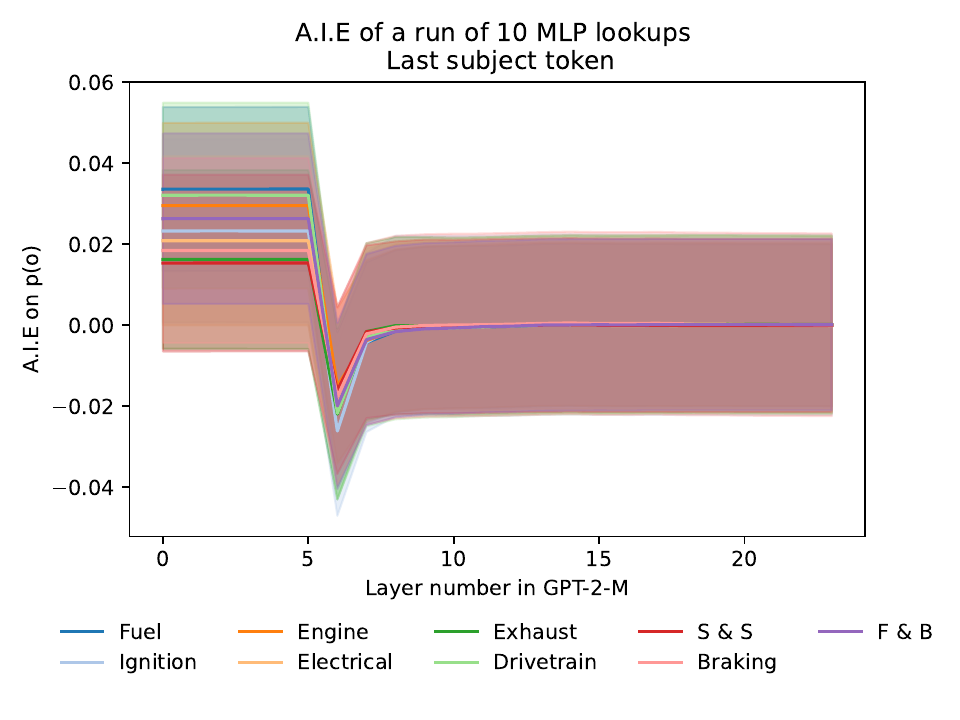} 
    \end{minipage}\hfill
    \begin{minipage}{0.48\textwidth}
        \centering
        \includegraphics[width=1.0\textwidth]{Images/m/Last_subject_token_mlp_body.pdf} 
    \end{minipage}

\caption{GPT-2-M Line Plots - Last Token - Attention }
\label{fig:GPT2M_Lineplots_Attn}
\end{figure}

\begin{figure*}[]
  \centering
  \centerline{\includegraphics[width=\linewidth]{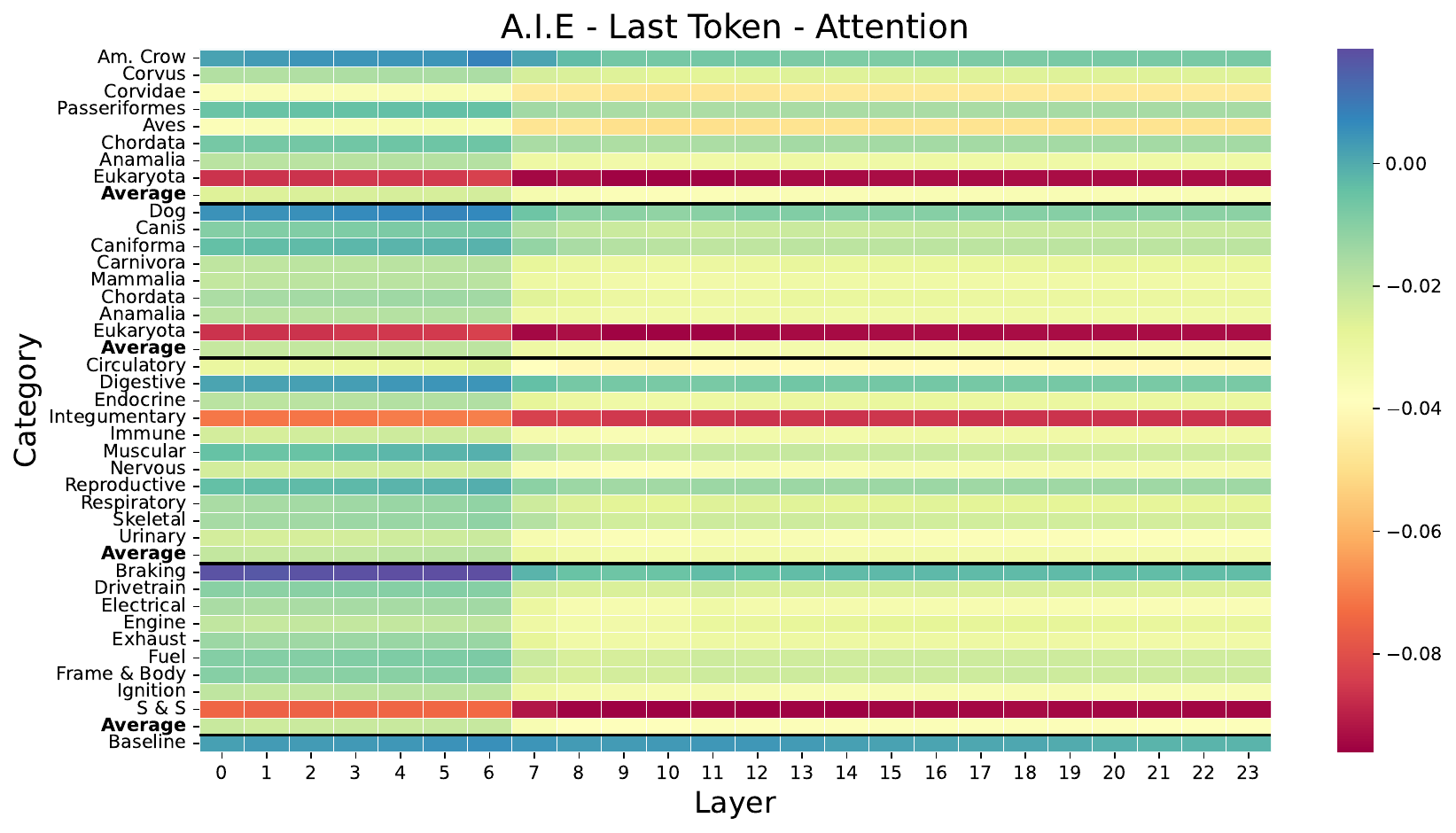}}

\caption{GPT-2-M A.I.E Heat Map - Attention}
\label{fig:LIH-M-Attn}

\end{figure*}

\end{document}